%% file: paper.tex
\documentclass{article}
\pdfoutput=1

\usepackage[final]{neurips_2021_ml4ad}

\usepackage[utf8]{inputenc} % allow utf-8 input
\usepackage[T1]{fontenc}    % use 8-bit T1 fonts
\usepackage{hyperref}       % hyperlinks
\usepackage{url}            % simple URL typesetting
\usepackage{booktabs}       % professional-quality tables
\usepackage{amsfonts}       % blackboard math symbols
\usepackage{nicefrac}       % compact symbols for 1/2, etc.
\usepackage{microtype}      % microtypography
\usepackage{xcolor}         % colors
\usepackage{pgf}
\setcitestyle{numbers,square}
\usepackage{layouts}

\title{Compressing Sensor Data for Remote Assistance of Autonomous Vehicles using Deep Generative Models}
\author{%
  Daniel Bogdoll$^{1}$\textsuperscript{\textasteriskcentered} \\
  \And
  Johannes Jestram$^{2}$\textsuperscript{\textasteriskcentered} \\
  \And
  Jonas Rauch$^{2}$\textsuperscript{\textasteriskcentered} \\
  \AND
  Christin Scheib$^{2}$\textsuperscript{\textasteriskcentered} \\
  \And
  Moritz Wittig$^{2}$\textsuperscript{\textasteriskcentered} \\
  \And
  J. Marius Zöllner$^{1}$
}

\begin{document}
%\printinunitsof{in}\prntlen{\textwidth} % [JJ] TODO Delete this later

\maketitle

\footnotetext[1]{FZI Research Center for Information Technology, Germany; {\tt\small \{bogdoll, zoellner\}@fzi.de}}
\footnotetext[2]{Karlsruhe Institute of Technology, Germany; {\tt\small \{johannes.jestram, jonas.rauch, christin.scheib, moritz.wittig\}@student.kit.edu}}
\begingroup\renewcommand\thefootnote{\textasteriskcentered}
\footnotetext{These authors contributed equally}
\endgroup

\input{sections/abstract}

\input{sections/introduction}
\input{sections/related-work}

\input{sections/approach}
\input{sections/evaluation}
\input{sections/conclusion}
\input{sections/acknowledgment}

\bibliography{bibliography}

%%%%%%%%%%%%%%%%%%%%%%%%%%%%%%%%%%%%%%%%%%%%%%%%%%%%%%%%%%%%

\input{sections/appendix.tex}

\end{document}

%% file: sections/abstract.tex
\begin{abstract}

In the foreseeable future, autonomous vehicles will require human assistance in situations they can not resolve on their own. In such scenarios, remote assistance from a human can provide the required input for the vehicle to continue its operation. Typical sensors used in autonomous vehicles include camera and lidar sensors. Due to the massive volume of sensor data that must be sent in real-time, highly efficient data compression is elementary to prevent an overload of network infrastructure. Sensor data compression using deep generative neural networks has been shown to outperform traditional compression approaches for both image and lidar data, regarding compression rate as well as reconstruction quality. However, there is a lack of research about the performance of generative-neural-network-based compression algorithms for remote assistance. In order to gain insights into the feasibility of deep generative models for usage in remote assistance, we evaluate state-of-the-art algorithms regarding their applicability and identify potential weaknesses. Further, we implement an online pipeline for processing sensor data and demonstrate its performance for remote assistance using the CARLA simulator.

\end{abstract}

%% file: sections/introduction.tex
\section{Introduction}
\label{sec:introduction}
\textbf{Motivation.} Both SAE International (SAE) level 4 and 5 autonomous vehicles (AV)~\cite{sae_2018} operate without relying on a human driver. However, in such complex environments as road traffic, AVs will not be able to operate without failures in the foreseeable future. Remote assistance~\cite{bogdoll2021taxonomy} allows remote operators to resolve situations, where AVs face problems they cannot resolve on their own by granting a remote operator access to the vehicle's state and sensor data~\cite{georg_adaptable_2019}. Depending on the specific situation and the implementation of the remote assistance, the remote operators can either provide the vehicle information that enable it to continue the operation, or take over manual control. 

Thus, to not overload network infrastructures, the volume of data that is sent must be drastically reduced. To achieve this, efficient compression of sensor data is elementary. Recent research on deep generative neural networks has achieved impressive results in image and lidar scan compression~\cite{caccia_deep_2019,Agustsson_2019, Mentzer_2020, Balle_2018, Cheng_2020}. Deep generative models are an attractive choice as they work with diverse sensor data and can achieve higher compression rates while maintaining a better reconstruction quality than JPEG and MPEG~\cite{Balle_2018}.
% In contrast, compression formats such as JPEG do not work well with sensor data from stereo cameras, which are often found in AVs~\cite{theis_lossy_2017}.
Further, such models also work for lidar scans and outperform tree-based and JPEG-based approaches~\cite{tu_compressing_2016, caccia_deep_2019, tu_point_2019}. Lastly, deep generative models allow combining encoders and decoders of different depths with each other, making them well suited for applications with differing hardware capabilities on the AV and the remote operator side~\cite{theis_lossy_2017}.

\textbf{Gap to related work.} Existing research focuses on compressing either image or lidar data~\cite{caccia_deep_2019, tu_compressing_2016, tu_point_2019, Agustsson_2019, Mentzer_2020, Balle_2018, Cheng_2020}. However, there is a lack of studies regarding the choice of compression approaches for the remote assistance of AVs. Multiple approaches exist that compare generative models with traditional codecs~\cite{Loehdefink_2019, Dash_2020}. Instead of comparing just one generative approach to traditional engineered algorithms for image compression, we provide a comprehensive comparison in the field of autonomous driving between multiple generative models and traditional algorithms.

As point cloud and image data are complementary, a single compression approach for processing both sensor modalities in the application of remote assistance is beneficial but has not been demonstrated yet in the literature. Further, to the best of our knowledge, there is no study that empirically evaluates the reduction in volume of data that can be achieved by employing compressing a vehicle's sensor data using deep generative models.

\textbf{Contributions.}
Therefore, our main motivation is to identify suitable approaches for compressing sensor data for remote assistance of autonomous vehicles. We thoroughly evaluate two state-of-the-art image compression approaches regarding their fit for remote assistance, using different datasets from the autonomous driving domain.

Based on several error metrics for reconstruction, we identify scenarios where generative compression performs either very good, or very bad. Additionally, we evaluate the reconstruction quality by performing object detection on the original as well as the reconstructed images.

We demonstrate a distributed online-pipeline for processing simulated image and lidar data, and evaluate its performance for our use case. Our chosen architecture allows camera and lidar data to be processed by the same pipeline, rather than having to perform individual compression passes.

We study the performance of the online pipeline with regards to its real-time capabilities, and highlight critical processing steps that affect performance. By implementing the online processing pipeline in the Robot Operating System (ROS)~\cite{ros}, we show the applicability of our approach on our test vehicle~\cite{zofka_2015_data}, that operates on the same framework.

\textbf{Paper outline.} The rest of this paper is structured as follows. Section~\ref{sec:related-work} introduces generative-model-based approaches for image and lidar compression and applications of compression pipelines. Section~\ref{sec:approach} explains our model choice and implementation of the online pipeline. Finally, we evaluate the image and lidar compression approaches in Section~\ref{sec:evaluation}, before concluding our work in Section~\ref{sec:conclusion}.

%% file: sections/related-work.tex
\section{Related Work}
\label{sec:related-work}
\subsection{Approaches for Image Compression}
\label{subsec:abbroaches_for_image_compression}

Traditional, lossy image compression algorithms, such as JPEG~\cite{Wallace_1991}, JPEG2000~\cite{taubmann_jpeg2000_2002} or HEVC~\cite{Sullivan_2012} based BPG~\cite{bellard_fabrice_bpg_2021}, are popular and commonly known. Nevertheless, neural networks have also been used for image compression for more than 20 years~\cite{Jiang_1999}. In 2006 Hinton and Salakhutdinov~\cite{Hinton_2006} introduced a deep autoencoder to convert high-dimensional data to low-dimensional codes. One key problem of the standard autoencoder is that it generates a fixed-length code for  images with the same resolution, independent of the complexity. Toderici et al.~\cite{Toderici_2015} introduced a variable-rate image compression framework where LSTM models are used for both the encoder and the decoder. The results showed remarkable improvements compared to JPEG on the similarity metric SSIM. In recent years Variational Autoencoders (VAE) and Generative Adversarial Networks (GAN) have reached a lot of attention due to their success in the field of image compression. GANs produce high quality, perceptual reconstructions, but with the danger of mode collapse. The  reconstructed images by VAEs are often more blurry and not necessarily as visually appealing to humans, but due to their setup mode-collapse is not an issue. 

Ball\'e et al.~\cite{Balle_2018} introduced a VAE for image compression with an hierarchical prior to improve the entropy model. The data compressed by the prior can be used as side information for the compression of the latent representation with the entropy model. Thereby, the entropy model can adjust to different complexities of images. Based on this, Minnen et al.~\cite{Minnen_2018} extended the hyperprior to a mean and scale Gaussian distribution alongside an autoregressive component that predicts latents from their causal context to get a more accurate entropy model. Cheng et al.~\cite{Cheng_2020} further improved the model using discretized Gaussian Mixture Likelihoods to parameterize the distributions of latent codes.

GANs~\cite{Goodfellow_2014} have led to impressive results for learning intractable distributions with generative models. Santurkar et al.~\cite{Santurkar_18} show how GANs can be used for lossy image compression by adding an autoencoder to a GAN architecture.

Adversarial losses are also used in the field of image compression in the rate-distortion objective~\cite{Mentzer_2018, Blau_2019}. Agustsson et al.~\cite{Agustsson_2019} show impressive, subjective results in their user study with extremely high compression rates. Mentzer et al.~\cite{Mentzer_2020} improve the distortion quality by the introduction of the hyperprior approach on the latents of Ball\'e et al.~\cite{Balle_2018}. The inherent danger of  mode collapse for GANs is tackled by the additional distortion loss, which  optimizes the reconstructions on the pixel level and thereby penalizes a mode collapse.

\subsection{Point Cloud Compression}
Several directions for compressing 3D data, including lidar measurements, have been explored. Ochotta and Saupe~\cite{ochotta_image-based_2008} propose an approach that decomposes dense 3D surfaces into smaller patches represented as elevation maps and applies a shape-adaptive wavelet encoder on those patches. For dense point clouds, tree-based compression approaches have been proposed~\cite{Schnabel2006OctreebasedPC}.~\citeauthor{Golla2015RealTime}~\cite{Golla2015RealTime} achieve real-time compression of point clouds by grouping points into larger voxels and applying compression on each voxel separately. They compute height-, color-, and occupancy-maps and compress those maps using different standard compression approaches, such as JPEG. However, lidar-generated point clouds tend to be sparsely populated. Therefore, the aforementioned approaches are not well-suited for our use case.

Tu~et~al.~\cite{tu_compressing_2016} suggest representing raw lidar scans as 2D matrices. They propose utilizing the knowledge that most lidars generate scans by rotating an array of lasers by exactly one revolution. This means that there is an inherent order to the array raw sensor data that the lidar produces, although it might differ between manufacturers. Utilizing this order allows the 2D matrix representation to be created relatively cheaply. To this 2D representation of the raw lidar data Tu~et~al. subsequently apply compression approaches, such as JPEG and MPEG~\cite{tu_compressing_2016}, an RNN with residual blocks~\cite{tu_point_2019}, and U-Net for optical flow interpolation between reference frames~\cite{tu_real_time_2019}. However, the RNN-based approach performs compression by sending data through the encoder as well as the decoder, making it not suitable for a distributed use case such as remote assistance. 

Another recent line of work explores point cloud compression using deep generative models, such as VAEs and GANs. After transformation of raw lidar scans into 2D grids, represented as matrices~\cite{tu_compressing_2016}, the encoder of a CNN-based VAE or GAN can compress the matrix analogous to an image. Conditional generation, i.e., reconstruction of a compressed scan, as well as compression rate have been shown to perform well with a VAE using such an ordered 2D representation~\cite{caccia_deep_2019}.

\subsection{Generative Data Compression Applications}

There exist several publications on generative models for image compression. Regarding comparative studies Löhdefink~et~al.~\cite{Loehdefink_2019} implement the GAN approach of~\cite{Agustsson_2019} and compare it to JPEG2000 with similarity metrics. Dash et al.~\cite{Dash_2020} improve the GAN architecture and compare the results to traditional compression algorithms. Siam et al.~\cite{Siam_2018} evaluate different approaches for image compression in the autonomous driving domain on semantically segmented images.

Outside of the autonomous driving community, end-to-end image compression based on generative models has been applied successfully to various other applications, such as facial images for surveillance~\cite{He_2018} and general video compression~\cite{Lu_2019, Habibian_2019}.

%% file: sections/approach.tex
\section{Approach}
\label{sec:approach}

Our work is divided into two parts, an offline and an online segment. In the offline part we select two state-of-the-art generative models for image and lidar compression to evaluate the compression and quality of the reconstructions and compare them to traditional approaches. We implement these approaches in an online AV simulation to evaluate their real-world potential. Code and supplementary material is available at \url{https://github.com/daniel-bogdoll/deep_generative_models}.

\textbf{Offline Image Compression.}\label{subsec:offline_image_compression}
For remote assistance of AVs efficient data compression with a high quality reconstruction is necessary. Therefore, our objective is to evaluate models based on their rate-distortion-perception \cite{blau2019rethinking} properties. We include the perception category, since the receiving remote operator is human, and assess it based on a domain-specific object-detection metric.

To meet these mentioned requirements we select the VAE model of Ball\'e~et~al.~\cite{Balle_2018} and the GAN model by Mentzer~et~al.~\cite{Mentzer_2020} as they achieved state-of-the-art results in the rate-distortion for image compression. Both models take a $256 \times 256$ pixels resize of the original image as an input and have been validated on various datasets and metrics.

\textbf{Offline Point Cloud Compression.}
\label{subsec:offline_point_cloud_compression_approach}
To use a VAE or GAN for lidar compression, we pre-process the data according to Caccia~et~al.~\cite{caccia_deep_2019}. The intuition behind the pre-processing approach is to sort and down-sample the lidar scans into a tensor representation, so that the resulting structure closely resembles the structure of an RGB image. We further base our compression of lidar scans on a 2D transformation and use a VAE for compression.  %This implies that points that are neighbors according to their euclidean distance should also reside close to each other in the tensor. Therefore, the resulting tensor $\mathcal{PC}$ has three dimensions, namely rows, columns, and one channel containing $(x,y,z)$-values. With the sensor as origin, the rows and columns correspond to the polar angle $\theta$ of each point and its azimuth angle $\phi$, respectively. Figure~\ref{fig:point_cloud_transformed} shows the format of the tensor after pre-processing.
We perform compression on diverse lidar data sources, i.e., KITTI~\cite{geiger_are_2012}, Waymo Open Perception Dataset~\cite{sun2020scalability}, and CARLA~\cite{dosovitskiy_carla_2017}. 

\textbf{Online Compression.}
To evaluate and compare the performance of the aforementioned approaches with regard to their applicability in remote assistance systems, we design an online sensor data compression pipeline. We use ROS to interconnect the different sub-systems. Furthermore, we employ the CARLA Simulator as our sensor data source. CARLA is an open-source simulator for development, training, and validation of autonomous urban driving systems~\cite{dosovitskiy_carla_2017}. A schematic overview of the system is given in Figure~\ref{fig:online_pipeline}. Examples of the CARLA environment and the decoded camera image are shown in the Appendix, Figure~\ref{fig:CARLA_VAE_compression}.

\begin{figure*}
    \includegraphics[width=\textwidth]{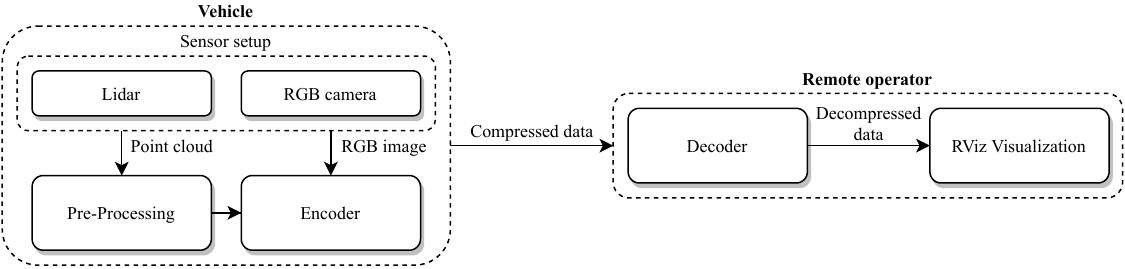}
    \caption{Overview of the online processing pipeline. The pre-processed lidar data passes through the same encoder and decoder architecture as RGB data.}
\label{fig:online_pipeline}
\end{figure*}

The AV in the CARLA simulator is equipped with a set of sensors such as a lidar and camera. The system consists of an encoder node on the AV side and a decoder node on the remote operator side. In case of compressing lidar data, an additional pre-processing node on the vehicle side is necessary. The sensor data is sent to the respective encoder node to compress the data. Afterwards, the decoder receives the data via a network connection and decompresses it.

%% file: sections/evaluation.tex
\section{Evaluation}
\label{sec:evaluation}
\subsection{Image Compression}
\textbf{Training.}
\label{subsubsec:training_images}
Models of the methods selected in Section~\ref{sec:approach} were trained using the KITTI \cite{geiger_are_2012} dataset, which consists of 7,481 RGB training images and 7,518 RGB test images with a resolution of $1242 \times 375$ pixels. 
During training, fragments with a size of $256 \times 256$ pixels were randomly cut out of the training images and fed to the corresponding model. Using this input size makes a good compromise between quality and compression rate.
The VAE was trained with a batch size of 32 and a learning rate of 0.0001 for 100 epochs to achieve good convergence of validation metric values. 
The library compressai \cite{begaint2020compressai} was used to perform training. 
Rate-distortion-loss, consisting of MSE as distortion loss and bit-rate as rate loss, was used as a combined loss function for the VAE. It is calculated as follows:

\begin{equation} 
    \mathcal{L}_{V} = \mathcal{D}_{V} + \lambda \mathcal{R}_{V} = \mathcal{L}_{MSE} + \lambda \mathcal{L}_{bpp}.
\end{equation}

The GAN was trained for 8 epochs using a learning rate of 0.0001 and a batch size of 4. 
As distortion loss, learned perceptual image patch similarity (LPIPS) \cite{Zhang2018} was used additionally. 
Therefore, the loss function used is calculated by
\begin{equation} 
    \mathcal{L}_{G} = \mathcal{D}_{G} + \lambda \mathcal{R}_{G} = ( k_m \mathcal{L}_{MSE} + k_p \mathcal{L}_{LPIPS} ) + \lambda \mathcal{L}_{bpp}
\end{equation}
where $k_m = 0.075 \cdot 2^{-5}$ and $k_p=1$ are hyper-parameters chosen as in \cite{Mentzer_2020}. 
Several trainings with different values of $\lambda \in \{ 0.001, 0.0025, 0.05, 0.01, 0.05, 0.1\}$ were performed to achieve different quality levels in reconstruction and different compression rates. 
We executed all trainings on an NVIDIA GeForce GTX 1080 Ti. 

\begin{figure}[t]
    \begin{center}
        \input{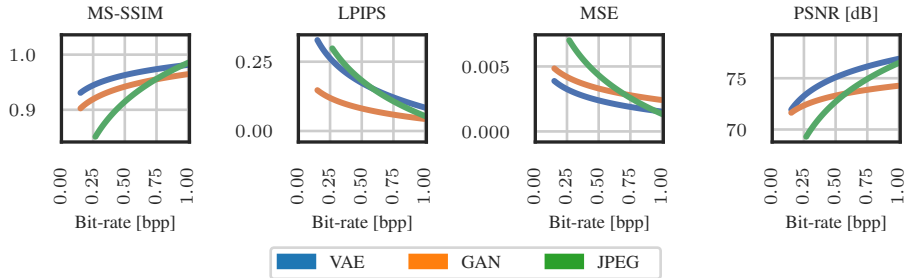}
    \end{center}
    \caption{Comparison of VAE, GAN and JPEG2000 compression with the metrics MS-SSIM, LPIPS, MSE and PSNR. The performances are plotted over the bit-rate in bits per pixel.}
    \label{vae_gan_jpeg_metrics_comparision}
\end{figure}

\textbf{Image Reconstruction Quality.}
\label{subsubsec:image_reconstruction_quality}
We compare the methods at different quality levels and show what level of compression still provides sufficient quality for the use case of remote assistance. 
For the evaluation of the image reconstruction quality, the metrics MSE, LPIPS, peak signal to noise ratio (PSNR) and multiscale structural similarity index measure (MS-SSIM) \cite{Wang2003} are taken into account. Both VAE and GAN image reconstruction results are evaluated and compared to JPEG2000, see Section~\ref{subsec:abbroaches_for_image_compression}. 
Figure \ref{vae_gan_jpeg_metrics_comparision} presents the metric values over bit-rates from 0 to 1.0 bits per pixel (bpp).

The JPEG compression delivers competitive results, but gets worse at lower bit-rates. 
This becomes especially clear at bit-rates below 0.3~bpp, where JPEG fails completely, since compression rates become unusable. However, we downsize every image to $256 \times 256$ pixels as done for VAE and GAN approaches, which negatively influences the reconstruction quality. 
For structural similarity, VAE and GAN yield similar trends. For MSE and PSNR, GAN and VAE approaches perform the same for lower bit-rates.\\ At higher bit-rates, the curves diverge. For LPIPS the GAN approach performs much better. 
Nevertheless, it should be noted here that the GAN was trained on LPIPS in addition to MSE in distribution loss which impacts the evaluation. To demonstrate the differences, multiple examples are presented.

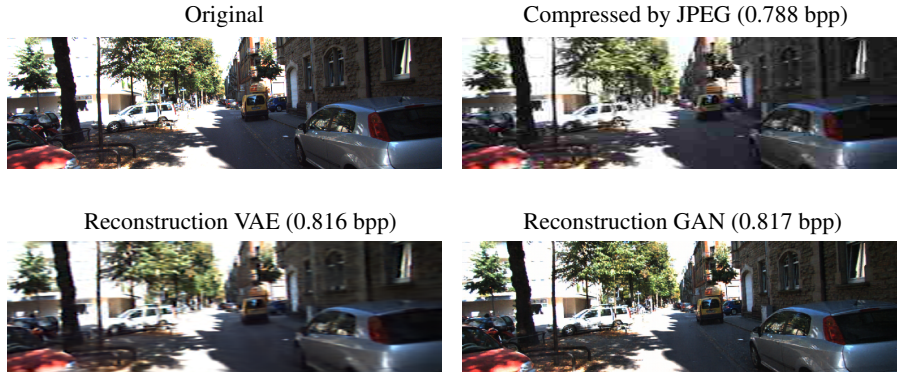
\begin{figure}[t]
    \begin{center}
        \input{figures/Compare-Compression-with-JPEG-Same-BPP.pgf}
    \end{center}
    \caption{KITTI street scene with reconstruction comparisons. Target bit-rate is approx. $0.8\,bpp$.}
    \label{Compare_Compression_with_JPEG_Same_BPP}
\end{figure}

First, Figure~\ref{Compare_Compression_with_JPEG_Same_BPP} displays a standard street scene from the KITTI dataset. It shows the original image, the JPEG compression and the reconstructions made by VAE and GAN approaches. 
In order to compare the results at the same compression rates, a similar target bit-rate was chosen for all compressions leading to equally large compressed data.

JPEG achieves the worst compression quality at these bit-rates. Details are hard to discern and the image looks fragmented. With the VAE approach, the reconstruction looks better, but more washed out. 
The GAN provides the best reconstruction, also showing details in higher sharpness. This can be seen with the tree in the background and with the gap between the right car's trunk and its side parts.

Further, two out-of-domain corner case situations, as described in \cite{bogdoll2021description},  were used as input for the KITTI trained models to test the stability of the approaches. The first case is a night scene from the Nighttime Driving dataset \cite{Dai2018} and the second case is a scene from the Fishyscapes benchmark \cite{Blum2021} that has an unconventional object positioned in a street scene. Figure \ref{out_of_domain_figure} in the appendix presents the results. Note that the target bit-rate is set lower than the one in Figure~\ref{Compare_Compression_with_JPEG_Same_BPP} to point out the methods' performances with lower bit-rates. While both generative approaches produce good results, poor performance can be seen with JPEG compression. At low bit-rates, colors may not be represented correctly, and there are also large fragments in the image. In contrast, the GAN approach still gives very good results, even for out-of-domain scenes for which it has not been trained specifically. Based on this, JPEG2000 was dismissed from further evaluation since GAN and VAE provide high-quality results with lower bit-rates than JPEG2000 can handle.

\begin{figure}
    \begin{center}
        \input{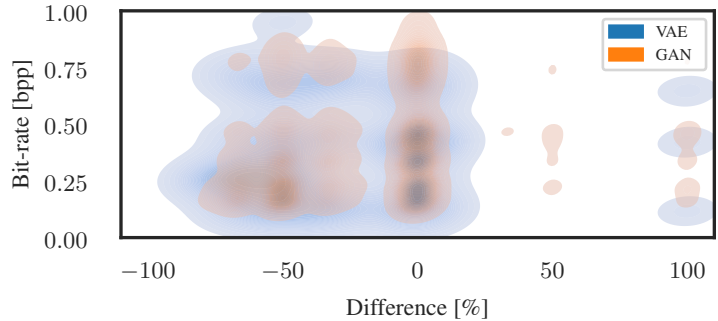}
    \end{center}
    \caption{Relative error of the number of detected cars in the original image and the reconstructed image by VAE respectively GAN. The distribution is given over the bit-rate in bpp. Scott's rule \cite{kde_scott} was used for smoothing with a scaling factor of 0.6. The darker the color in the graph, the higher the density of the values.}
    \label{object_detection_kde}
\end{figure}

\textbf{Object Detection Performance in Reconstructions.}
While the metric evaluation performed in Section~\ref{subsubsec:image_reconstruction_quality}  assesses the overall quality of reconstructions, the following evaluation refers to the remote operator's requirement, which is to understand the scene. Therefore, the extent to which object recognition is possible on the reconstructed image is investigated. For this purpose, an SSD \cite{Liu2016} model with ResNet50 \cite{He2016} as feature extractor trained on COCO \cite{Lin2014} was used. Object detection for the class car was performed on both the original and the reconstructed images. For evaluation, the number of cars detected in the original and the reconstructed image were counted, followed by a comparison using the following formula for the relative error:

\begin{equation}
    \mathit{Relative \, error} = \frac{n_{\mathit{recon}} - n_{\mathit{orig}}}{n_{\mathit{orig}}} \times 100
\end{equation}

where $n_{\mathit{recon}}$ is the number of cars detected in the reconstructed image and $n_{\mathit{orig}}$ in the original image. Figure \ref{object_detection_kde} shows this deviation in percent, i.e., -100\% deviation means that out of $n$ cars in the original image, none were detected in the reconstructed image. Images with $n_{orig}=0$ were not considered. The minimum confidence score for the object detection network was chosen as $0.7$.  It is noticeable that across all quality levels, the deviations in the negative range are large, especially for VAE. Also, there is a large number of images where the GAN approach has almost no errors. Likewise, there are some images reconstructed by both VAE or GAN for which the object detection algorithm detects more objects as in the original counterparts, with the maximum difference being 100\%. Overall, images reconstructed with the GAN allow for better object recognition of cars.

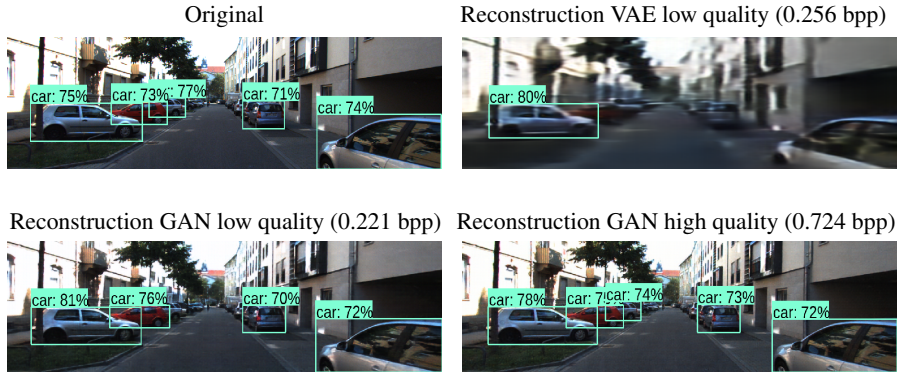
\begin{figure}
    \begin{center}
        \input{figures/object_detection_compare.pgf}
    \end{center}
    \caption{Object detection example performed in the original image and several reconstructions made by GAN and VAE with differing bit-rates.}
    \label{object_detection_compare}
\end{figure}

Figure \ref{object_detection_compare} shows an example of the object detection performance. The VAE compression with low quality performs inadequately with the pre-trained object detection model. However, at the same bit-rate the GAN compression results in a better object detection, improving with a higher bit-rate.

Considering that objects in the distance might be of less importance, even lower bit-rates in the range of 0.2 to 0.3~bpp are possible with the GAN. At such low rates, the VAE compression performs insufficient. Therefore, using a GAN trained with low $\lambda$ values is preferable and achieves high-quality reconstructions with a high compression. 

\subsection{Point Cloud Compression}
Early performance tests, which are further elaborated in Section~\ref{subsec:online_pipeline}, have shown the significantly higher transmission rates of VAEs. Therefore, we focus on VAEs for point cloud compression. The combined inference time for both camera and lidar GAN processing would be too high for the targeted real-world application. The VAE approach introduced in Section~\ref{sec:approach} was trained and tested on KITTI lidar data. The same training parameters as in Section~\ref{subsubsec:training_images} were used. As input the VAE receives the preprocessed point clouds with a shape of $512 \times 64$, since a general resizing of the point clouds as it is done for images is not possible as the channels contain distances and not color value information. Due to the different information type, the bit-rate increases as well.

\begin{figure}
    \centering
    \includegraphics[width=0.6\linewidth]{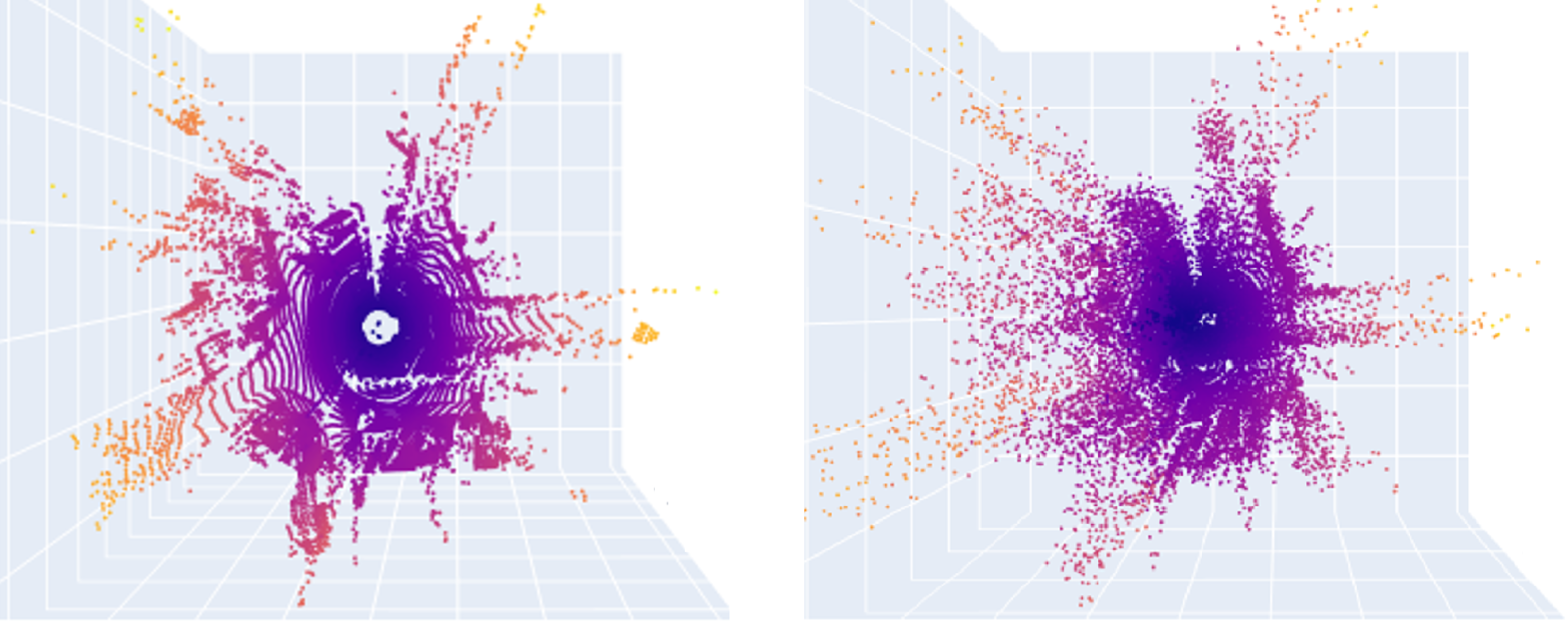}
    \caption{VAE point cloud reconstruction. Left: Original, transformed point cloud. Right: Reconstruction with a bit-rate of 1.83~bpp. The scenery is viewed from above.}
    \label{fig:point_cloud_example}
\end{figure}

Despite using the same VAE architecture as used for image compression, the results in compressing and reconstructing point clouds are surprisingly good. We measure the point-wise reconstruction quality with the standard euclidean distance metric. For a target bit-rate of 1.8~bpp, the mean euclidean distance is between $0.3\,\mathrm{m}$ and $0.5\,\mathrm{m}$, while points close to the center have a smaller distance to their counterparts than points located further away.

Figure~\ref{fig:point_cloud_example} shows that the typical lidar rings are no longer reconstructed, and that there is increased noise. This can also be seen in the PSNR value, which is about 48.0 at a target bit-rate of 1.8~bpp. Still, general distance trends can be identified and certain deflections can be recognized. Therefore, this approach generates early, but promising results.

\subsection{Online Pipeline}
\label{subsec:online_pipeline}

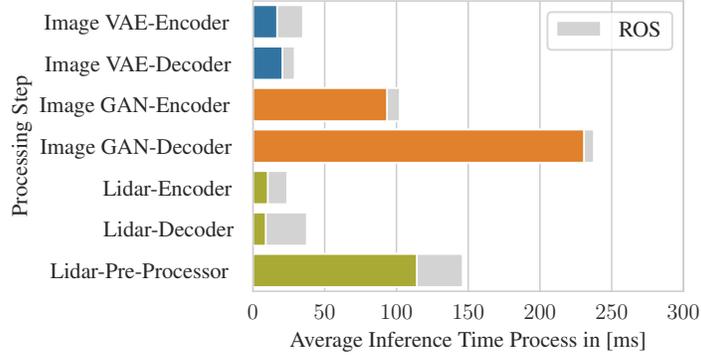
\begin{figure}
    \begin{center}
        \scalebox{0.7}{\input{figures/online-bar-inference-with-ROS-overhead.pgf}}
    \end{center}
    \caption{Average inference time for the different ROS nodes.}
\label{fig:online-bar-inference-with-ROS-overhead.pgf}
\end{figure}

In a remote assistance system the transmission of the sensor data is most important for taking appropriate action. As the sensor data must be encoded and decoded to increase data throughput, the transmission of the sensor data is prone to latency. Therefore, fast processing times are necessary. We test the compression approaches introduced in Section~\ref{sec:approach} regarding their processing times and throughput on an NVIDIA GeForce GTX 1080 Ti to investigate their suitability for the use case of remote assistance. We measure the processing time of the nodes over a time span of 300~seconds. The processing time gives information about the throughput of the pre-processor, encoder and decoder. To understand the overhead introduced by ROS we also measure the mere inference time for only encoding and decoding of the images and point clouds. The results are summarized in  Figure~\ref{fig:online-bar-inference-with-ROS-overhead.pgf}. 

For image compression the VAE requires significantly less processing time than the GAN, leading to a throughput of about 28~FPS. The ROS encoder has a mean processing time of 34.9~ms of which 16.8~ms is the mere inference time for the VAE compression. The ROS decoding node requires 29.1~ms while the mere decompression takes 20.5~ms. The resulting system latency adds up to 57.1~ms due to the processing times of encoder and decoder in addition to the network latency. The generation of the ROS message from the latent representation and its publishing is more time consuming than the image compression itself. The decoder must first reconstruct the latent representation from the transmitted ROS message which is less time consuming.  

In the GAN-based image compression pipeline the processing times of the encoder and decoder vary significantly. While the ROS encoder has a mean processing time of 102.5~ms, the complete pipeline can only process 4~FPS, as the decoder requires about 237.7~ms per frame. This results in a latency of about 340.2~ms. Without the ROS overhead the inference time for the image compression is about 93.4~ms for the encoder and 230.8~ms for the decoder. Comparing this to the VAE the relative overhead introduced by ROS is significantly smaller. This can be traced back to a time-consuming process for the generation of the customized ROS message, representing the VAE's latent space. 

For point cloud compression the pre-processing node in its current implementation is a bottleneck. It determines the throughput of 6~scans per second of the complete pipeline. The mean processing time for the pre-processing node is 146.5~ms while the encoder and decoder node reach processing times of 23.9~ms and 37.7~ms respectively. Investigating the mere inference time without ROS overhead, the discrepancy between the lidar pre-processing and its compression itself is significant as well. Pre-processing already takes 114.5~ms per scan while compression only takes~10.4 ms and decompression~8.9 ms.

While the GAN and the point cloud compression reach impressive compression results, their processing times are too high to be currently applied in a remote assistance system. Georg~et~al.~\cite{latency} consider frame rates of about 30~FPS sufficient for remote driving in low speed scenarios while for high speed scenarios a frame rate of 55~FPS is needed. For remote driving latency values above 300 ms make the safe operation of the AV almost impossible~\cite{neumeier_feasibility_2019}. While the latency requirements for remote assistance~\cite{space_teleop} are less stringent compared to remote driving~\cite{trajectory_control}, an operator still requires real-time knowledge of the state of the vehicle's environment to make decisions.

%% file: figures/Compare-Compression-with-JPEG-Same-BPP.pgf
%% Creator: Matplotlib, PGF backend
%%
%% To include the figure in your LaTeX document, write
%%   \input{<filename>.pgf}
%%
%% Make sure the required packages are loaded in your preamble
%%   \usepackage{pgf}
%%
%% Figures using additional raster images can only be included by \input if
%% they are in the same directory as the main LaTeX file. For loading figures
%% from other directories you can use the `import` package
%%   \usepackage{import}
%% and then include the figures with
%%   \import{<path to file>}{<filename>.pgf}
%%
%% Matplotlib used the following preamble
%%
\begingroup%
\makeatletter%
\begin{pgfpicture}%
\pgfpathrectangle{\pgfpointorigin}{\pgfqpoint{4.850000in}{2.231806in}}%
\pgfusepath{use as bounding box, clip}%
\begin{pgfscope}%
\pgfsetbuttcap%
\pgfsetmiterjoin%
\definecolor{currentfill}{rgb}{1.000000,1.000000,1.000000}%
\pgfsetfillcolor{currentfill}%
\pgfsetlinewidth{0.000000pt}%
\definecolor{currentstroke}{rgb}{1.000000,1.000000,1.000000}%
\pgfsetstrokecolor{currentstroke}%
\pgfsetdash{}{0pt}%
\pgfpathmoveto{\pgfqpoint{0.000000in}{0.000000in}}%
\pgfpathlineto{\pgfqpoint{4.850000in}{0.000000in}}%
\pgfpathlineto{\pgfqpoint{4.850000in}{2.231806in}}%
\pgfpathlineto{\pgfqpoint{0.000000in}{2.231806in}}%
\pgfpathclose%
\pgfusepath{fill}%
\end{pgfscope}%
\begin{pgfscope}%
\pgfpathrectangle{\pgfqpoint{0.100000in}{1.165455in}}{\pgfqpoint{2.268293in}{0.684871in}}%
\pgfusepath{clip}%
\pgfsys@transformshift{0.100000in}{1.165455in}%
\pgftext[left,bottom]{\pgfimage[interpolate=true,width=2.270000in,height=0.685000in]{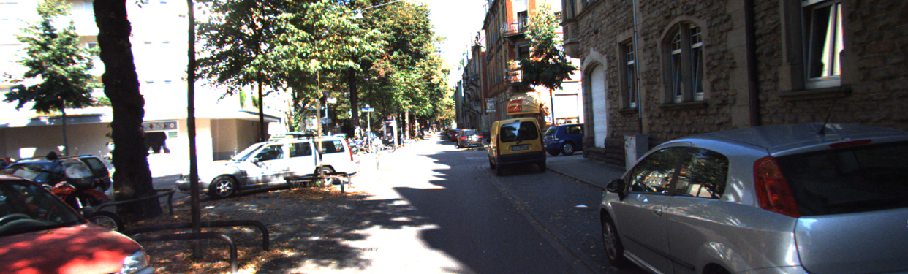}}%
\end{pgfscope}%
\begin{pgfscope}%
\definecolor{textcolor}{rgb}{0.000000,0.000000,0.000000}%
\pgfsetstrokecolor{textcolor}%
\pgfsetfillcolor{textcolor}%
\pgftext[x=1.234146in,y=1.933in,,base]{\color{textcolor}\rmfamily\fontsize{9.000000}{14.400000}\selectfont Original}%
\end{pgfscope}%
\begin{pgfscope}%
\pgfpathrectangle{\pgfqpoint{2.481707in}{1.165455in}}{\pgfqpoint{2.268293in}{0.684871in}}%
\pgfusepath{clip}%
\pgfsys@transformshift{2.481707in}{1.165455in}%
\pgftext[left,bottom]{\pgfimage[interpolate=true,width=2.270000in,height=0.685000in]{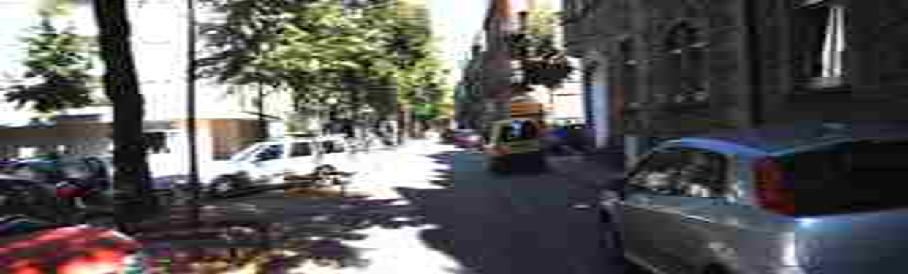}}%
\end{pgfscope}%
\begin{pgfscope}%
\definecolor{textcolor}{rgb}{0.000000,0.000000,0.000000}%
\pgfsetstrokecolor{textcolor}%
\pgfsetfillcolor{textcolor}%
\pgftext[x=2.8in,y=1.933in,left,base]{\color{textcolor}\rmfamily\fontsize{9.000000}{14.400000}\selectfont Compressed by JPEG (0.788 bpp)}%
\end{pgfscope}%
\begin{pgfscope}%
\definecolor{textcolor}{rgb}{0.000000,0.000000,0.000000}%
\pgfsetstrokecolor{textcolor}%
\pgfsetfillcolor{textcolor}%
\end{pgfscope}%
\begin{pgfscope}%
\pgfpathrectangle{\pgfqpoint{0.100000in}{0.100000in}}{\pgfqpoint{2.268293in}{0.684871in}}%
\pgfusepath{clip}%
\pgfsys@transformshift{0.100000in}{0.100000in}%
\pgftext[left,bottom]{\pgfimage[interpolate=true,width=2.270000in,height=0.685000in]{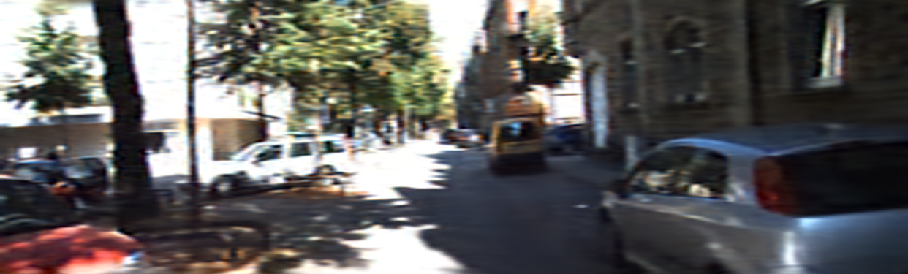}}%
\end{pgfscope}%
\begin{pgfscope}%
\definecolor{textcolor}{rgb}{0.000000,0.000000,0.000000}%
\pgfsetstrokecolor{textcolor}%
\pgfsetfillcolor{textcolor}%
\pgftext[x=0.5in,y=0.850611in,left,base]{\color{textcolor}\rmfamily\fontsize{9.000000}{14.400000}\selectfont Reconstruction VAE (0.816 bpp)}%
\end{pgfscope}%
\begin{pgfscope}%
\definecolor{textcolor}{rgb}{0.000000,0.000000,0.000000}%
\pgfsetstrokecolor{textcolor}%
\pgfsetfillcolor{textcolor}%
\end{pgfscope}%
\begin{pgfscope}%
\pgfpathrectangle{\pgfqpoint{2.481707in}{0.100000in}}{\pgfqpoint{2.268293in}{0.684871in}}%
\pgfusepath{clip}%
\pgfsys@transformshift{2.481707in}{0.100000in}%
\pgftext[left,bottom]{\pgfimage[interpolate=true,width=2.270000in,height=0.685000in]{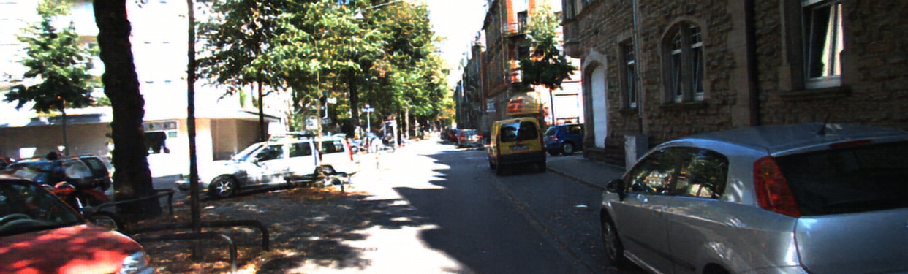}}%
\end{pgfscope}%
\begin{pgfscope}%
\definecolor{textcolor}{rgb}{0.000000,0.000000,0.000000}%
\pgfsetstrokecolor{textcolor}%
\pgfsetfillcolor{textcolor}%
\pgftext[x=2.8in,y=0.850611in,left,base]{\color{textcolor}\rmfamily\fontsize{9.000000}{14.400000}\selectfont Reconstruction GAN (0.817 bpp)}%
\end{pgfscope}%
\begin{pgfscope}%
\definecolor{textcolor}{rgb}{0.000000,0.000000,0.000000}%
\pgfsetstrokecolor{textcolor}%
\pgfsetfillcolor{textcolor}%
\end{pgfscope}%
\end{pgfpicture}%
\makeatother%
\endgroup%

%% file: figures/object_detection_compare.pgf
%% Creator: Matplotlib, PGF backend
%%
%% To include the figure in your LaTeX document, write
%%   \input{<filename>.pgf}
%%
%% Make sure the required packages are loaded in your preamble
%%   \usepackage{pgf}
%%
%% Figures using additional raster images can only be included by \input if
%% they are in the same directory as the main LaTeX file. For loading figures
%% from other directories you can use the `import` package
%%   \usepackage{import}
%% and then include the figures with
%%   \import{<path to file>}{<filename>.pgf}
%%
%% Matplotlib used the following preamble
%%
\begingroup%
\makeatletter%
\begin{pgfpicture}%
\pgfpathrectangle{\pgfpointorigin}{\pgfqpoint{4.850000in}{2.231806in}}%
\pgfusepath{use as bounding box, clip}%
\begin{pgfscope}%
\pgfsetbuttcap%
\pgfsetmiterjoin%
\definecolor{currentfill}{rgb}{1.000000,1.000000,1.000000}%
\pgfsetfillcolor{currentfill}%
\pgfsetlinewidth{0.000000pt}%
\definecolor{currentstroke}{rgb}{1.000000,1.000000,1.000000}%
\pgfsetstrokecolor{currentstroke}%
\pgfsetdash{}{0pt}%
\pgfpathmoveto{\pgfqpoint{0.000000in}{0.000000in}}%
\pgfpathlineto{\pgfqpoint{4.850000in}{0.000000in}}%
\pgfpathlineto{\pgfqpoint{4.850000in}{2.231806in}}%
\pgfpathlineto{\pgfqpoint{0.000000in}{2.231806in}}%
\pgfpathclose%
\pgfusepath{fill}%
\end{pgfscope}%
\begin{pgfscope}%
\pgfpathrectangle{\pgfqpoint{0.100000in}{1.165455in}}{\pgfqpoint{2.268293in}{0.684871in}}%
\pgfusepath{clip}%
\pgfsys@transformshift{0.100000in}{1.165455in}%
\pgftext[left,bottom]{\pgfimage[interpolate=true,width=2.270000in,height=0.685000in]{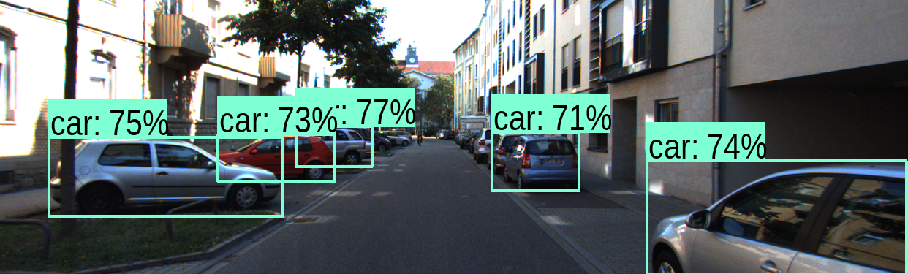}}%
\end{pgfscope}%
\begin{pgfscope}%
\definecolor{textcolor}{rgb}{0.000000,0.000000,0.000000}%
\pgfsetstrokecolor{textcolor}%
\pgfsetfillcolor{textcolor}%
\pgftext[x=1.234146in,y=1.933in,,base]{\color{textcolor}\rmfamily\fontsize{9.000000}{14.400000}\selectfont Original}%
\end{pgfscope}%
\begin{pgfscope}%
\pgfpathrectangle{\pgfqpoint{2.481707in}{1.165455in}}{\pgfqpoint{2.268293in}{0.684871in}}%
\pgfusepath{clip}%
\pgfsys@transformshift{2.481707in}{1.165455in}%
\pgftext[left,bottom]{\pgfimage[interpolate=true,width=2.270000in,height=0.685000in]{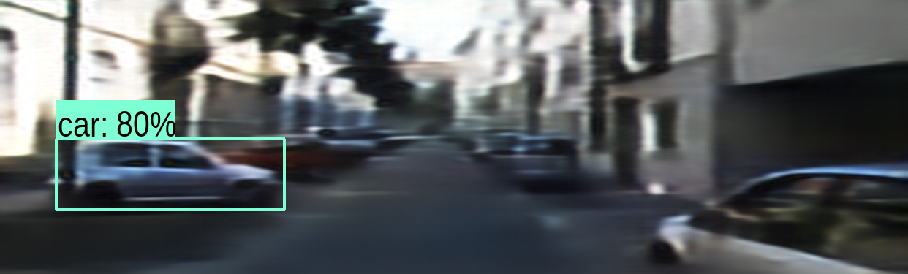}}%
\end{pgfscope}%
\begin{pgfscope}%
\definecolor{textcolor}{rgb}{0.000000,0.000000,0.000000}%
\pgfsetstrokecolor{textcolor}%
\pgfsetfillcolor{textcolor}%
\pgftext[x=2.47in,y=1.933in,left,base]{\color{textcolor}\rmfamily\fontsize{9.000000}{14.400000}\selectfont Reconstruction VAE low quality (0.256 bpp)}%
\end{pgfscope}%
\begin{pgfscope}%
\definecolor{textcolor}{rgb}{0.000000,0.000000,0.000000}%
\pgfsetstrokecolor{textcolor}%
\pgfsetfillcolor{textcolor}%
\end{pgfscope}%
\begin{pgfscope}%
\pgfpathrectangle{\pgfqpoint{0.100000in}{0.100000in}}{\pgfqpoint{2.268293in}{0.684871in}}%
\pgfusepath{clip}%
\pgfsys@transformshift{0.100000in}{0.100000in}%
\pgftext[left,bottom]{\pgfimage[interpolate=true,width=2.270000in,height=0.685000in]{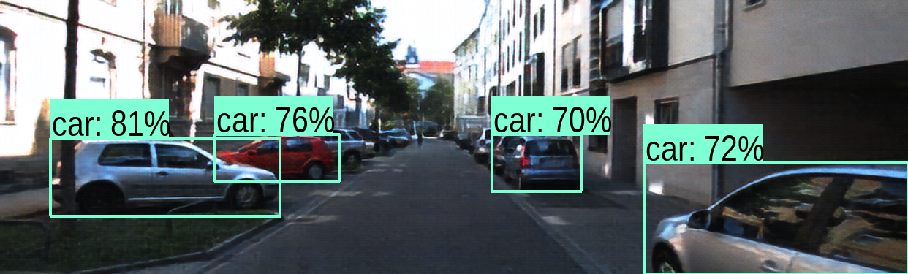}}%
\end{pgfscope}%
\begin{pgfscope}%
\definecolor{textcolor}{rgb}{0.000000,0.000000,0.000000}%
\pgfsetstrokecolor{textcolor}%
\pgfsetfillcolor{textcolor}%
\pgftext[x=0.11in,y=0.850611in,left,base]{\color{textcolor}\rmfamily\fontsize{9.000000}{14.400000}\selectfont Reconstruction GAN low quality (0.221 bpp)}%
\end{pgfscope}%
\begin{pgfscope}%
\definecolor{textcolor}{rgb}{0.000000,0.000000,0.000000}%
\pgfsetstrokecolor{textcolor}%
\pgfsetfillcolor{textcolor}%
\end{pgfscope}%
\begin{pgfscope}%
\pgfpathrectangle{\pgfqpoint{2.481707in}{0.100000in}}{\pgfqpoint{2.268293in}{0.684871in}}%
\pgfusepath{clip}%
\pgfsys@transformshift{2.481707in}{0.100000in}%
\pgftext[left,bottom]{\pgfimage[interpolate=true,width=2.270000in,height=0.685000in]{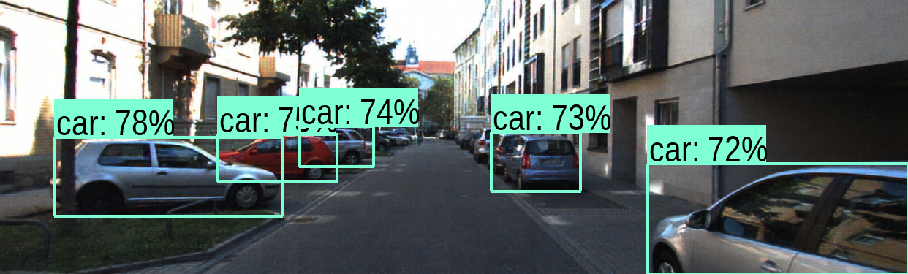}}%
\end{pgfscope}%
\begin{pgfscope}%
\definecolor{textcolor}{rgb}{0.000000,0.000000,0.000000}%
\pgfsetstrokecolor{textcolor}%
\pgfsetfillcolor{textcolor}%
\pgftext[x=2.45in,y=0.850611in,left,base]{\color{textcolor}\rmfamily\fontsize{9.000000}{14.400000}\selectfont Reconstruction GAN high quality (0.724 bpp)}%
\end{pgfscope}%
\begin{pgfscope}%
\definecolor{textcolor}{rgb}{0.000000,0.000000,0.000000}%
\pgfsetstrokecolor{textcolor}%
\pgfsetfillcolor{textcolor}%
\end{pgfscope}%
\end{pgfpicture}%
\makeatother%
\endgroup%

%% file: figures/online-bar-inference-with-ROS-overhead.pgf
%% Creator: Matplotlib, PGF backend
%%
%% To include the figure in your LaTeX document, write
%%   \input{<filename>.pgf}
%%
%% Make sure the required packages are loaded in your preamble
%%   \usepackage{pgf}
%%
%% and, on pdftex
%%   \usepackage[utf8]{inputenc}\DeclareUnicodeCharacter{2212}{-}
%%
%% or, on luatex and xetex
%%   \usepackage{unicode-math}
%%
%% Figures using additional raster images can only be included by \input if
%% they are in the same directory as the main LaTeX file. For loading figures
%% from other directories you can use the `import` package
%%   \usepackage{import}
%%
%% and then include the figures with
%%   \import{<path to file>}{<filename>.pgf}
%%
%% Matplotlib used the following preamble
%%
\begingroup%
\makeatletter%
\begin{pgfpicture}%
\pgfpathrectangle{\pgfpointorigin}{\pgfqpoint{5.501070in}{3.000000in}}%
\pgfusepath{use as bounding box, clip}%
\begin{pgfscope}%
\pgfsetbuttcap%
\pgfsetmiterjoin%
\definecolor{currentfill}{rgb}{1.000000,1.000000,1.000000}%
\pgfsetfillcolor{currentfill}%
\pgfsetlinewidth{0.000000pt}%
\definecolor{currentstroke}{rgb}{1.000000,1.000000,1.000000}%
\pgfsetstrokecolor{currentstroke}%
\pgfsetdash{}{0pt}%
\pgfpathmoveto{\pgfqpoint{0.000000in}{0.000000in}}%
\pgfpathlineto{\pgfqpoint{5.501070in}{0.000000in}}%
\pgfpathlineto{\pgfqpoint{5.501070in}{3.000000in}}%
\pgfpathlineto{\pgfqpoint{0.000000in}{3.000000in}}%
\pgfpathclose%
\pgfusepath{fill}%
\end{pgfscope}%
\begin{pgfscope}%
\pgfsetbuttcap%
\pgfsetmiterjoin%
\definecolor{currentfill}{rgb}{1.000000,1.000000,1.000000}%
\pgfsetfillcolor{currentfill}%
\pgfsetlinewidth{0.000000pt}%
\definecolor{currentstroke}{rgb}{0.000000,0.000000,0.000000}%
\pgfsetstrokecolor{currentstroke}%
\pgfsetstrokeopacity{0.000000}%
\pgfsetdash{}{0pt}%
\pgfpathmoveto{\pgfqpoint{1.980042in}{0.647592in}}%
\pgfpathlineto{\pgfqpoint{5.198675in}{0.647592in}}%
\pgfpathlineto{\pgfqpoint{5.198675in}{2.820000in}}%
\pgfpathlineto{\pgfqpoint{1.980042in}{2.820000in}}%
\pgfpathclose%
\pgfusepath{fill}%
\end{pgfscope}%
\begin{pgfscope}%
\pgfpathrectangle{\pgfqpoint{1.980042in}{0.647592in}}{\pgfqpoint{3.218633in}{2.172408in}}%
\pgfusepath{clip}%
\pgfsetroundcap%
\pgfsetroundjoin%
\pgfsetlinewidth{0.803000pt}%
\definecolor{currentstroke}{rgb}{0.800000,0.800000,0.800000}%
\pgfsetstrokecolor{currentstroke}%
\pgfsetdash{}{0pt}%
\pgfpathmoveto{\pgfqpoint{1.980042in}{0.647592in}}%
\pgfpathlineto{\pgfqpoint{1.980042in}{2.820000in}}%
\pgfusepath{stroke}%
\end{pgfscope}%
\begin{pgfscope}%
\definecolor{textcolor}{rgb}{0.150000,0.150000,0.150000}%
\pgfsetstrokecolor{textcolor}%
\pgfsetfillcolor{textcolor}%
\pgftext[x=1.980042in,y=0.550370in,,top]{\color{textcolor}\rmfamily\fontsize{12.000000}{14.400000}\selectfont \(\displaystyle {0}\)}%
\end{pgfscope}%
\begin{pgfscope}%
\pgfpathrectangle{\pgfqpoint{1.980042in}{0.647592in}}{\pgfqpoint{3.218633in}{2.172408in}}%
\pgfusepath{clip}%
\pgfsetroundcap%
\pgfsetroundjoin%
\pgfsetlinewidth{0.803000pt}%
\definecolor{currentstroke}{rgb}{0.800000,0.800000,0.800000}%
\pgfsetstrokecolor{currentstroke}%
\pgfsetdash{}{0pt}%
\pgfpathmoveto{\pgfqpoint{2.516481in}{0.647592in}}%
\pgfpathlineto{\pgfqpoint{2.516481in}{2.820000in}}%
\pgfusepath{stroke}%
\end{pgfscope}%
\begin{pgfscope}%
\definecolor{textcolor}{rgb}{0.150000,0.150000,0.150000}%
\pgfsetstrokecolor{textcolor}%
\pgfsetfillcolor{textcolor}%
\pgftext[x=2.516481in,y=0.550370in,,top]{\color{textcolor}\rmfamily\fontsize{12.000000}{14.400000}\selectfont \(\displaystyle {50}\)}%
\end{pgfscope}%
\begin{pgfscope}%
\pgfpathrectangle{\pgfqpoint{1.980042in}{0.647592in}}{\pgfqpoint{3.218633in}{2.172408in}}%
\pgfusepath{clip}%
\pgfsetroundcap%
\pgfsetroundjoin%
\pgfsetlinewidth{0.803000pt}%
\definecolor{currentstroke}{rgb}{0.800000,0.800000,0.800000}%
\pgfsetstrokecolor{currentstroke}%
\pgfsetdash{}{0pt}%
\pgfpathmoveto{\pgfqpoint{3.052920in}{0.647592in}}%
\pgfpathlineto{\pgfqpoint{3.052920in}{2.820000in}}%
\pgfusepath{stroke}%
\end{pgfscope}%
\begin{pgfscope}%
\definecolor{textcolor}{rgb}{0.150000,0.150000,0.150000}%
\pgfsetstrokecolor{textcolor}%
\pgfsetfillcolor{textcolor}%
\pgftext[x=3.052920in,y=0.550370in,,top]{\color{textcolor}\rmfamily\fontsize{12.000000}{14.400000}\selectfont \(\displaystyle {100}\)}%
\end{pgfscope}%
\begin{pgfscope}%
\pgfpathrectangle{\pgfqpoint{1.980042in}{0.647592in}}{\pgfqpoint{3.218633in}{2.172408in}}%
\pgfusepath{clip}%
\pgfsetroundcap%
\pgfsetroundjoin%
\pgfsetlinewidth{0.803000pt}%
\definecolor{currentstroke}{rgb}{0.800000,0.800000,0.800000}%
\pgfsetstrokecolor{currentstroke}%
\pgfsetdash{}{0pt}%
\pgfpathmoveto{\pgfqpoint{3.589359in}{0.647592in}}%
\pgfpathlineto{\pgfqpoint{3.589359in}{2.820000in}}%
\pgfusepath{stroke}%
\end{pgfscope}%
\begin{pgfscope}%
\definecolor{textcolor}{rgb}{0.150000,0.150000,0.150000}%
\pgfsetstrokecolor{textcolor}%
\pgfsetfillcolor{textcolor}%
\pgftext[x=3.589359in,y=0.550370in,,top]{\color{textcolor}\rmfamily\fontsize{12.000000}{14.400000}\selectfont \(\displaystyle {150}\)}%
\end{pgfscope}%
\begin{pgfscope}%
\pgfpathrectangle{\pgfqpoint{1.980042in}{0.647592in}}{\pgfqpoint{3.218633in}{2.172408in}}%
\pgfusepath{clip}%
\pgfsetroundcap%
\pgfsetroundjoin%
\pgfsetlinewidth{0.803000pt}%
\definecolor{currentstroke}{rgb}{0.800000,0.800000,0.800000}%
\pgfsetstrokecolor{currentstroke}%
\pgfsetdash{}{0pt}%
\pgfpathmoveto{\pgfqpoint{4.125798in}{0.647592in}}%
\pgfpathlineto{\pgfqpoint{4.125798in}{2.820000in}}%
\pgfusepath{stroke}%
\end{pgfscope}%
\begin{pgfscope}%
\definecolor{textcolor}{rgb}{0.150000,0.150000,0.150000}%
\pgfsetstrokecolor{textcolor}%
\pgfsetfillcolor{textcolor}%
\pgftext[x=4.125798in,y=0.550370in,,top]{\color{textcolor}\rmfamily\fontsize{12.000000}{14.400000}\selectfont \(\displaystyle {200}\)}%
\end{pgfscope}%
\begin{pgfscope}%
\pgfpathrectangle{\pgfqpoint{1.980042in}{0.647592in}}{\pgfqpoint{3.218633in}{2.172408in}}%
\pgfusepath{clip}%
\pgfsetroundcap%
\pgfsetroundjoin%
\pgfsetlinewidth{0.803000pt}%
\definecolor{currentstroke}{rgb}{0.800000,0.800000,0.800000}%
\pgfsetstrokecolor{currentstroke}%
\pgfsetdash{}{0pt}%
\pgfpathmoveto{\pgfqpoint{4.662237in}{0.647592in}}%
\pgfpathlineto{\pgfqpoint{4.662237in}{2.820000in}}%
\pgfusepath{stroke}%
\end{pgfscope}%
\begin{pgfscope}%
\definecolor{textcolor}{rgb}{0.150000,0.150000,0.150000}%
\pgfsetstrokecolor{textcolor}%
\pgfsetfillcolor{textcolor}%
\pgftext[x=4.662237in,y=0.550370in,,top]{\color{textcolor}\rmfamily\fontsize{12.000000}{14.400000}\selectfont \(\displaystyle {250}\)}%
\end{pgfscope}%
\begin{pgfscope}%
\pgfpathrectangle{\pgfqpoint{1.980042in}{0.647592in}}{\pgfqpoint{3.218633in}{2.172408in}}%
\pgfusepath{clip}%
\pgfsetroundcap%
\pgfsetroundjoin%
\pgfsetlinewidth{0.803000pt}%
\definecolor{currentstroke}{rgb}{0.800000,0.800000,0.800000}%
\pgfsetstrokecolor{currentstroke}%
\pgfsetdash{}{0pt}%
\pgfpathmoveto{\pgfqpoint{5.198675in}{0.647592in}}%
\pgfpathlineto{\pgfqpoint{5.198675in}{2.820000in}}%
\pgfusepath{stroke}%
\end{pgfscope}%
\begin{pgfscope}%
\definecolor{textcolor}{rgb}{0.150000,0.150000,0.150000}%
\pgfsetstrokecolor{textcolor}%
\pgfsetfillcolor{textcolor}%
\pgftext[x=5.198675in,y=0.550370in,,top]{\color{textcolor}\rmfamily\fontsize{12.000000}{14.400000}\selectfont \(\displaystyle {300}\)}%
\end{pgfscope}%
\begin{pgfscope}%
\definecolor{textcolor}{rgb}{0.150000,0.150000,0.150000}%
\pgfsetstrokecolor{textcolor}%
\pgfsetfillcolor{textcolor}%
\pgftext[x=3.589359in,y=0.346667in,,top]{\color{textcolor}\rmfamily\fontsize{12.000000}{14.400000}\selectfont Average Inference Time Process in [ms]}%
\end{pgfscope}%
\begin{pgfscope}%
\definecolor{textcolor}{rgb}{0.150000,0.150000,0.150000}%
\pgfsetstrokecolor{textcolor}%
\pgfsetfillcolor{textcolor}%
\pgftext[x=0.418849in, y=2.606958in, left, base]{\color{textcolor}\rmfamily\fontsize{12.000000}{14.400000}\selectfont Image VAE-Encoder}%
\end{pgfscope}%
\begin{pgfscope}%
\definecolor{textcolor}{rgb}{0.150000,0.150000,0.150000}%
\pgfsetstrokecolor{textcolor}%
\pgfsetfillcolor{textcolor}%
\pgftext[x=0.423383in, y=2.296614in, left, base]{\color{textcolor}\rmfamily\fontsize{12.000000}{14.400000}\selectfont Image VAE-Decoder}%
\end{pgfscope}%
\begin{pgfscope}%
\definecolor{textcolor}{rgb}{0.150000,0.150000,0.150000}%
\pgfsetstrokecolor{textcolor}%
\pgfsetfillcolor{textcolor}%
\pgftext[x=0.383703in, y=1.986270in, left, base]{\color{textcolor}\rmfamily\fontsize{12.000000}{14.400000}\selectfont Image GAN-Encoder}%
\end{pgfscope}%
\begin{pgfscope}%
\definecolor{textcolor}{rgb}{0.150000,0.150000,0.150000}%
\pgfsetstrokecolor{textcolor}%
\pgfsetfillcolor{textcolor}%
\pgftext[x=0.388236in, y=1.675926in, left, base]{\color{textcolor}\rmfamily\fontsize{12.000000}{14.400000}\selectfont Image GAN-Decoder}%
\end{pgfscope}%
\begin{pgfscope}%
\definecolor{textcolor}{rgb}{0.150000,0.150000,0.150000}%
\pgfsetstrokecolor{textcolor}%
\pgfsetfillcolor{textcolor}%
\pgftext[x=0.858390in, y=1.365582in, left, base]{\color{textcolor}\rmfamily\fontsize{12.000000}{14.400000}\selectfont Lidar-Encoder}%
\end{pgfscope}%
\begin{pgfscope}%
\definecolor{textcolor}{rgb}{0.150000,0.150000,0.150000}%
\pgfsetstrokecolor{textcolor}%
\pgfsetfillcolor{textcolor}%
\pgftext[x=0.862923in, y=1.055238in, left, base]{\color{textcolor}\rmfamily\fontsize{12.000000}{14.400000}\selectfont Lidar-Decoder}%
\end{pgfscope}%
\begin{pgfscope}%
\definecolor{textcolor}{rgb}{0.150000,0.150000,0.150000}%
\pgfsetstrokecolor{textcolor}%
\pgfsetfillcolor{textcolor}%
\pgftext[x=0.464489in, y=0.744894in, left, base]{\color{textcolor}\rmfamily\fontsize{12.000000}{14.400000}\selectfont Lidar-Pre-Processor}%
\end{pgfscope}%
\begin{pgfscope}%
\definecolor{textcolor}{rgb}{0.150000,0.150000,0.150000}%
\pgfsetstrokecolor{textcolor}%
\pgfsetfillcolor{textcolor}%
\pgftext[x=0.328148in,y=1.733796in,,bottom,rotate=90.000000]{\color{textcolor}\rmfamily\fontsize{12.000000}{14.400000}\selectfont Processing Step}%
\end{pgfscope}%
\begin{pgfscope}%
\pgfpathrectangle{\pgfqpoint{1.980042in}{0.647592in}}{\pgfqpoint{3.218633in}{2.172408in}}%
\pgfusepath{clip}%
\pgfsetbuttcap%
\pgfsetmiterjoin%
\definecolor{currentfill}{rgb}{0.827451,0.827451,0.827451}%
\pgfsetfillcolor{currentfill}%
\pgfsetlinewidth{1.003750pt}%
\definecolor{currentstroke}{rgb}{1.000000,1.000000,1.000000}%
\pgfsetstrokecolor{currentstroke}%
\pgfsetdash{}{0pt}%
\pgfpathmoveto{\pgfqpoint{1.980042in}{2.788966in}}%
\pgfpathlineto{\pgfqpoint{2.354413in}{2.788966in}}%
\pgfpathlineto{\pgfqpoint{2.354413in}{2.540690in}}%
\pgfpathlineto{\pgfqpoint{1.980042in}{2.540690in}}%
\pgfpathclose%
\pgfusepath{stroke,fill}%
\end{pgfscope}%
\begin{pgfscope}%
\pgfpathrectangle{\pgfqpoint{1.980042in}{0.647592in}}{\pgfqpoint{3.218633in}{2.172408in}}%
\pgfusepath{clip}%
\pgfsetbuttcap%
\pgfsetmiterjoin%
\definecolor{currentfill}{rgb}{0.827451,0.827451,0.827451}%
\pgfsetfillcolor{currentfill}%
\pgfsetlinewidth{1.003750pt}%
\definecolor{currentstroke}{rgb}{1.000000,1.000000,1.000000}%
\pgfsetstrokecolor{currentstroke}%
\pgfsetdash{}{0pt}%
\pgfpathmoveto{\pgfqpoint{1.980042in}{2.478622in}}%
\pgfpathlineto{\pgfqpoint{2.292669in}{2.478622in}}%
\pgfpathlineto{\pgfqpoint{2.292669in}{2.230346in}}%
\pgfpathlineto{\pgfqpoint{1.980042in}{2.230346in}}%
\pgfpathclose%
\pgfusepath{stroke,fill}%
\end{pgfscope}%
\begin{pgfscope}%
\pgfpathrectangle{\pgfqpoint{1.980042in}{0.647592in}}{\pgfqpoint{3.218633in}{2.172408in}}%
\pgfusepath{clip}%
\pgfsetbuttcap%
\pgfsetmiterjoin%
\definecolor{currentfill}{rgb}{0.827451,0.827451,0.827451}%
\pgfsetfillcolor{currentfill}%
\pgfsetlinewidth{1.003750pt}%
\definecolor{currentstroke}{rgb}{1.000000,1.000000,1.000000}%
\pgfsetstrokecolor{currentstroke}%
\pgfsetdash{}{0pt}%
\pgfpathmoveto{\pgfqpoint{1.980042in}{2.168278in}}%
\pgfpathlineto{\pgfqpoint{3.078503in}{2.168278in}}%
\pgfpathlineto{\pgfqpoint{3.078503in}{1.920002in}}%
\pgfpathlineto{\pgfqpoint{1.980042in}{1.920002in}}%
\pgfpathclose%
\pgfusepath{stroke,fill}%
\end{pgfscope}%
\begin{pgfscope}%
\pgfpathrectangle{\pgfqpoint{1.980042in}{0.647592in}}{\pgfqpoint{3.218633in}{2.172408in}}%
\pgfusepath{clip}%
\pgfsetbuttcap%
\pgfsetmiterjoin%
\definecolor{currentfill}{rgb}{0.827451,0.827451,0.827451}%
\pgfsetfillcolor{currentfill}%
\pgfsetlinewidth{1.003750pt}%
\definecolor{currentstroke}{rgb}{1.000000,1.000000,1.000000}%
\pgfsetstrokecolor{currentstroke}%
\pgfsetdash{}{0pt}%
\pgfpathmoveto{\pgfqpoint{1.980042in}{1.857934in}}%
\pgfpathlineto{\pgfqpoint{4.529529in}{1.857934in}}%
\pgfpathlineto{\pgfqpoint{4.529529in}{1.609658in}}%
\pgfpathlineto{\pgfqpoint{1.980042in}{1.609658in}}%
\pgfpathclose%
\pgfusepath{stroke,fill}%
\end{pgfscope}%
\begin{pgfscope}%
\pgfpathrectangle{\pgfqpoint{1.980042in}{0.647592in}}{\pgfqpoint{3.218633in}{2.172408in}}%
\pgfusepath{clip}%
\pgfsetbuttcap%
\pgfsetmiterjoin%
\definecolor{currentfill}{rgb}{0.827451,0.827451,0.827451}%
\pgfsetfillcolor{currentfill}%
\pgfsetlinewidth{1.003750pt}%
\definecolor{currentstroke}{rgb}{1.000000,1.000000,1.000000}%
\pgfsetstrokecolor{currentstroke}%
\pgfsetdash{}{0pt}%
\pgfpathmoveto{\pgfqpoint{1.980042in}{1.547590in}}%
\pgfpathlineto{\pgfqpoint{2.236347in}{1.547590in}}%
\pgfpathlineto{\pgfqpoint{2.236347in}{1.299314in}}%
\pgfpathlineto{\pgfqpoint{1.980042in}{1.299314in}}%
\pgfpathclose%
\pgfusepath{stroke,fill}%
\end{pgfscope}%
\begin{pgfscope}%
\pgfpathrectangle{\pgfqpoint{1.980042in}{0.647592in}}{\pgfqpoint{3.218633in}{2.172408in}}%
\pgfusepath{clip}%
\pgfsetbuttcap%
\pgfsetmiterjoin%
\definecolor{currentfill}{rgb}{0.827451,0.827451,0.827451}%
\pgfsetfillcolor{currentfill}%
\pgfsetlinewidth{1.003750pt}%
\definecolor{currentstroke}{rgb}{1.000000,1.000000,1.000000}%
\pgfsetstrokecolor{currentstroke}%
\pgfsetdash{}{0pt}%
\pgfpathmoveto{\pgfqpoint{1.980042in}{1.237246in}}%
\pgfpathlineto{\pgfqpoint{2.383446in}{1.237246in}}%
\pgfpathlineto{\pgfqpoint{2.383446in}{0.988970in}}%
\pgfpathlineto{\pgfqpoint{1.980042in}{0.988970in}}%
\pgfpathclose%
\pgfusepath{stroke,fill}%
\end{pgfscope}%
\begin{pgfscope}%
\pgfpathrectangle{\pgfqpoint{1.980042in}{0.647592in}}{\pgfqpoint{3.218633in}{2.172408in}}%
\pgfusepath{clip}%
\pgfsetbuttcap%
\pgfsetmiterjoin%
\definecolor{currentfill}{rgb}{0.827451,0.827451,0.827451}%
\pgfsetfillcolor{currentfill}%
\pgfsetlinewidth{1.003750pt}%
\definecolor{currentstroke}{rgb}{1.000000,1.000000,1.000000}%
\pgfsetstrokecolor{currentstroke}%
\pgfsetdash{}{0pt}%
\pgfpathmoveto{\pgfqpoint{1.980042in}{0.926902in}}%
\pgfpathlineto{\pgfqpoint{3.549042in}{0.926902in}}%
\pgfpathlineto{\pgfqpoint{3.549042in}{0.678626in}}%
\pgfpathlineto{\pgfqpoint{1.980042in}{0.678626in}}%
\pgfpathclose%
\pgfusepath{stroke,fill}%
\end{pgfscope}%
\begin{pgfscope}%
\pgfpathrectangle{\pgfqpoint{1.980042in}{0.647592in}}{\pgfqpoint{3.218633in}{2.172408in}}%
\pgfusepath{clip}%
\pgfsetbuttcap%
\pgfsetmiterjoin%
\definecolor{currentfill}{rgb}{0.194608,0.453431,0.632843}%
\pgfsetfillcolor{currentfill}%
\pgfsetlinewidth{1.003750pt}%
\definecolor{currentstroke}{rgb}{1.000000,1.000000,1.000000}%
\pgfsetstrokecolor{currentstroke}%
\pgfsetdash{}{0pt}%
\pgfpathmoveto{\pgfqpoint{1.980042in}{2.788966in}}%
\pgfpathlineto{\pgfqpoint{2.161107in}{2.788966in}}%
\pgfpathlineto{\pgfqpoint{2.161107in}{2.540690in}}%
\pgfpathlineto{\pgfqpoint{1.980042in}{2.540690in}}%
\pgfpathclose%
\pgfusepath{stroke,fill}%
\end{pgfscope}%
\begin{pgfscope}%
\pgfpathrectangle{\pgfqpoint{1.980042in}{0.647592in}}{\pgfqpoint{3.218633in}{2.172408in}}%
\pgfusepath{clip}%
\pgfsetbuttcap%
\pgfsetmiterjoin%
\definecolor{currentfill}{rgb}{0.194608,0.453431,0.632843}%
\pgfsetfillcolor{currentfill}%
\pgfsetlinewidth{1.003750pt}%
\definecolor{currentstroke}{rgb}{1.000000,1.000000,1.000000}%
\pgfsetstrokecolor{currentstroke}%
\pgfsetdash{}{0pt}%
\pgfpathmoveto{\pgfqpoint{1.980042in}{2.478622in}}%
\pgfpathlineto{\pgfqpoint{2.200459in}{2.478622in}}%
\pgfpathlineto{\pgfqpoint{2.200459in}{2.230346in}}%
\pgfpathlineto{\pgfqpoint{1.980042in}{2.230346in}}%
\pgfpathclose%
\pgfusepath{stroke,fill}%
\end{pgfscope}%
\begin{pgfscope}%
\pgfpathrectangle{\pgfqpoint{1.980042in}{0.647592in}}{\pgfqpoint{3.218633in}{2.172408in}}%
\pgfusepath{clip}%
\pgfsetbuttcap%
\pgfsetmiterjoin%
\definecolor{currentfill}{rgb}{0.881863,0.505392,0.173039}%
\pgfsetfillcolor{currentfill}%
\pgfsetlinewidth{1.003750pt}%
\definecolor{currentstroke}{rgb}{1.000000,1.000000,1.000000}%
\pgfsetstrokecolor{currentstroke}%
\pgfsetdash{}{0pt}%
\pgfpathmoveto{\pgfqpoint{1.980042in}{2.168278in}}%
\pgfpathlineto{\pgfqpoint{2.982394in}{2.168278in}}%
\pgfpathlineto{\pgfqpoint{2.982394in}{1.920002in}}%
\pgfpathlineto{\pgfqpoint{1.980042in}{1.920002in}}%
\pgfpathclose%
\pgfusepath{stroke,fill}%
\end{pgfscope}%
\begin{pgfscope}%
\pgfpathrectangle{\pgfqpoint{1.980042in}{0.647592in}}{\pgfqpoint{3.218633in}{2.172408in}}%
\pgfusepath{clip}%
\pgfsetbuttcap%
\pgfsetmiterjoin%
\definecolor{currentfill}{rgb}{0.881863,0.505392,0.173039}%
\pgfsetfillcolor{currentfill}%
\pgfsetlinewidth{1.003750pt}%
\definecolor{currentstroke}{rgb}{1.000000,1.000000,1.000000}%
\pgfsetstrokecolor{currentstroke}%
\pgfsetdash{}{0pt}%
\pgfpathmoveto{\pgfqpoint{1.980042in}{1.857934in}}%
\pgfpathlineto{\pgfqpoint{4.455409in}{1.857934in}}%
\pgfpathlineto{\pgfqpoint{4.455409in}{1.609658in}}%
\pgfpathlineto{\pgfqpoint{1.980042in}{1.609658in}}%
\pgfpathclose%
\pgfusepath{stroke,fill}%
\end{pgfscope}%
\begin{pgfscope}%
\pgfpathrectangle{\pgfqpoint{1.980042in}{0.647592in}}{\pgfqpoint{3.218633in}{2.172408in}}%
\pgfusepath{clip}%
\pgfsetbuttcap%
\pgfsetmiterjoin%
\definecolor{currentfill}{rgb}{0.662255,0.665196,0.209314}%
\pgfsetfillcolor{currentfill}%
\pgfsetlinewidth{1.003750pt}%
\definecolor{currentstroke}{rgb}{1.000000,1.000000,1.000000}%
\pgfsetstrokecolor{currentstroke}%
\pgfsetdash{}{0pt}%
\pgfpathmoveto{\pgfqpoint{1.980042in}{1.547590in}}%
\pgfpathlineto{\pgfqpoint{2.091094in}{1.547590in}}%
\pgfpathlineto{\pgfqpoint{2.091094in}{1.299314in}}%
\pgfpathlineto{\pgfqpoint{1.980042in}{1.299314in}}%
\pgfpathclose%
\pgfusepath{stroke,fill}%
\end{pgfscope}%
\begin{pgfscope}%
\pgfpathrectangle{\pgfqpoint{1.980042in}{0.647592in}}{\pgfqpoint{3.218633in}{2.172408in}}%
\pgfusepath{clip}%
\pgfsetbuttcap%
\pgfsetmiterjoin%
\definecolor{currentfill}{rgb}{0.662255,0.665196,0.209314}%
\pgfsetfillcolor{currentfill}%
\pgfsetlinewidth{1.003750pt}%
\definecolor{currentstroke}{rgb}{1.000000,1.000000,1.000000}%
\pgfsetstrokecolor{currentstroke}%
\pgfsetdash{}{0pt}%
\pgfpathmoveto{\pgfqpoint{1.980042in}{1.237246in}}%
\pgfpathlineto{\pgfqpoint{2.075679in}{1.237246in}}%
\pgfpathlineto{\pgfqpoint{2.075679in}{0.988970in}}%
\pgfpathlineto{\pgfqpoint{1.980042in}{0.988970in}}%
\pgfpathclose%
\pgfusepath{stroke,fill}%
\end{pgfscope}%
\begin{pgfscope}%
\pgfpathrectangle{\pgfqpoint{1.980042in}{0.647592in}}{\pgfqpoint{3.218633in}{2.172408in}}%
\pgfusepath{clip}%
\pgfsetbuttcap%
\pgfsetmiterjoin%
\definecolor{currentfill}{rgb}{0.662255,0.665196,0.209314}%
\pgfsetfillcolor{currentfill}%
\pgfsetlinewidth{1.003750pt}%
\definecolor{currentstroke}{rgb}{1.000000,1.000000,1.000000}%
\pgfsetstrokecolor{currentstroke}%
\pgfsetdash{}{0pt}%
\pgfpathmoveto{\pgfqpoint{1.980042in}{0.926902in}}%
\pgfpathlineto{\pgfqpoint{3.205436in}{0.926902in}}%
\pgfpathlineto{\pgfqpoint{3.205436in}{0.678626in}}%
\pgfpathlineto{\pgfqpoint{1.980042in}{0.678626in}}%
\pgfpathclose%
\pgfusepath{stroke,fill}%
\end{pgfscope}%
\begin{pgfscope}%
\pgfpathrectangle{\pgfqpoint{1.980042in}{0.647592in}}{\pgfqpoint{3.218633in}{2.172408in}}%
\pgfusepath{clip}%
\pgfsetroundcap%
\pgfsetroundjoin%
\pgfsetlinewidth{2.710125pt}%
\definecolor{currentstroke}{rgb}{0.260000,0.260000,0.260000}%
\pgfsetstrokecolor{currentstroke}%
\pgfsetdash{}{0pt}%
\pgfusepath{stroke}%
\end{pgfscope}%
\begin{pgfscope}%
\pgfpathrectangle{\pgfqpoint{1.980042in}{0.647592in}}{\pgfqpoint{3.218633in}{2.172408in}}%
\pgfusepath{clip}%
\pgfsetroundcap%
\pgfsetroundjoin%
\pgfsetlinewidth{2.710125pt}%
\definecolor{currentstroke}{rgb}{0.260000,0.260000,0.260000}%
\pgfsetstrokecolor{currentstroke}%
\pgfsetdash{}{0pt}%
\pgfusepath{stroke}%
\end{pgfscope}%
\begin{pgfscope}%
\pgfpathrectangle{\pgfqpoint{1.980042in}{0.647592in}}{\pgfqpoint{3.218633in}{2.172408in}}%
\pgfusepath{clip}%
\pgfsetroundcap%
\pgfsetroundjoin%
\pgfsetlinewidth{2.710125pt}%
\definecolor{currentstroke}{rgb}{0.260000,0.260000,0.260000}%
\pgfsetstrokecolor{currentstroke}%
\pgfsetdash{}{0pt}%
\pgfusepath{stroke}%
\end{pgfscope}%
\begin{pgfscope}%
\pgfpathrectangle{\pgfqpoint{1.980042in}{0.647592in}}{\pgfqpoint{3.218633in}{2.172408in}}%
\pgfusepath{clip}%
\pgfsetroundcap%
\pgfsetroundjoin%
\pgfsetlinewidth{2.710125pt}%
\definecolor{currentstroke}{rgb}{0.260000,0.260000,0.260000}%
\pgfsetstrokecolor{currentstroke}%
\pgfsetdash{}{0pt}%
\pgfusepath{stroke}%
\end{pgfscope}%
\begin{pgfscope}%
\pgfpathrectangle{\pgfqpoint{1.980042in}{0.647592in}}{\pgfqpoint{3.218633in}{2.172408in}}%
\pgfusepath{clip}%
\pgfsetroundcap%
\pgfsetroundjoin%
\pgfsetlinewidth{2.710125pt}%
\definecolor{currentstroke}{rgb}{0.260000,0.260000,0.260000}%
\pgfsetstrokecolor{currentstroke}%
\pgfsetdash{}{0pt}%
\pgfusepath{stroke}%
\end{pgfscope}%
\begin{pgfscope}%
\pgfpathrectangle{\pgfqpoint{1.980042in}{0.647592in}}{\pgfqpoint{3.218633in}{2.172408in}}%
\pgfusepath{clip}%
\pgfsetroundcap%
\pgfsetroundjoin%
\pgfsetlinewidth{2.710125pt}%
\definecolor{currentstroke}{rgb}{0.260000,0.260000,0.260000}%
\pgfsetstrokecolor{currentstroke}%
\pgfsetdash{}{0pt}%
\pgfusepath{stroke}%
\end{pgfscope}%
\begin{pgfscope}%
\pgfpathrectangle{\pgfqpoint{1.980042in}{0.647592in}}{\pgfqpoint{3.218633in}{2.172408in}}%
\pgfusepath{clip}%
\pgfsetroundcap%
\pgfsetroundjoin%
\pgfsetlinewidth{2.710125pt}%
\definecolor{currentstroke}{rgb}{0.260000,0.260000,0.260000}%
\pgfsetstrokecolor{currentstroke}%
\pgfsetdash{}{0pt}%
\pgfusepath{stroke}%
\end{pgfscope}%
\begin{pgfscope}%
\pgfpathrectangle{\pgfqpoint{1.980042in}{0.647592in}}{\pgfqpoint{3.218633in}{2.172408in}}%
\pgfusepath{clip}%
\pgfsetroundcap%
\pgfsetroundjoin%
\pgfsetlinewidth{2.710125pt}%
\definecolor{currentstroke}{rgb}{0.260000,0.260000,0.260000}%
\pgfsetstrokecolor{currentstroke}%
\pgfsetdash{}{0pt}%
\pgfusepath{stroke}%
\end{pgfscope}%
\begin{pgfscope}%
\pgfpathrectangle{\pgfqpoint{1.980042in}{0.647592in}}{\pgfqpoint{3.218633in}{2.172408in}}%
\pgfusepath{clip}%
\pgfsetroundcap%
\pgfsetroundjoin%
\pgfsetlinewidth{2.710125pt}%
\definecolor{currentstroke}{rgb}{0.260000,0.260000,0.260000}%
\pgfsetstrokecolor{currentstroke}%
\pgfsetdash{}{0pt}%
\pgfusepath{stroke}%
\end{pgfscope}%
\begin{pgfscope}%
\pgfpathrectangle{\pgfqpoint{1.980042in}{0.647592in}}{\pgfqpoint{3.218633in}{2.172408in}}%
\pgfusepath{clip}%
\pgfsetroundcap%
\pgfsetroundjoin%
\pgfsetlinewidth{2.710125pt}%
\definecolor{currentstroke}{rgb}{0.260000,0.260000,0.260000}%
\pgfsetstrokecolor{currentstroke}%
\pgfsetdash{}{0pt}%
\pgfusepath{stroke}%
\end{pgfscope}%
\begin{pgfscope}%
\pgfpathrectangle{\pgfqpoint{1.980042in}{0.647592in}}{\pgfqpoint{3.218633in}{2.172408in}}%
\pgfusepath{clip}%
\pgfsetroundcap%
\pgfsetroundjoin%
\pgfsetlinewidth{2.710125pt}%
\definecolor{currentstroke}{rgb}{0.260000,0.260000,0.260000}%
\pgfsetstrokecolor{currentstroke}%
\pgfsetdash{}{0pt}%
\pgfusepath{stroke}%
\end{pgfscope}%
\begin{pgfscope}%
\pgfpathrectangle{\pgfqpoint{1.980042in}{0.647592in}}{\pgfqpoint{3.218633in}{2.172408in}}%
\pgfusepath{clip}%
\pgfsetroundcap%
\pgfsetroundjoin%
\pgfsetlinewidth{2.710125pt}%
\definecolor{currentstroke}{rgb}{0.260000,0.260000,0.260000}%
\pgfsetstrokecolor{currentstroke}%
\pgfsetdash{}{0pt}%
\pgfusepath{stroke}%
\end{pgfscope}%
\begin{pgfscope}%
\pgfpathrectangle{\pgfqpoint{1.980042in}{0.647592in}}{\pgfqpoint{3.218633in}{2.172408in}}%
\pgfusepath{clip}%
\pgfsetroundcap%
\pgfsetroundjoin%
\pgfsetlinewidth{2.710125pt}%
\definecolor{currentstroke}{rgb}{0.260000,0.260000,0.260000}%
\pgfsetstrokecolor{currentstroke}%
\pgfsetdash{}{0pt}%
\pgfusepath{stroke}%
\end{pgfscope}%
\begin{pgfscope}%
\pgfpathrectangle{\pgfqpoint{1.980042in}{0.647592in}}{\pgfqpoint{3.218633in}{2.172408in}}%
\pgfusepath{clip}%
\pgfsetroundcap%
\pgfsetroundjoin%
\pgfsetlinewidth{2.710125pt}%
\definecolor{currentstroke}{rgb}{0.260000,0.260000,0.260000}%
\pgfsetstrokecolor{currentstroke}%
\pgfsetdash{}{0pt}%
\pgfusepath{stroke}%
\end{pgfscope}%
\begin{pgfscope}%
\pgfsetrectcap%
\pgfsetmiterjoin%
\pgfsetlinewidth{0.803000pt}%
\definecolor{currentstroke}{rgb}{0.800000,0.800000,0.800000}%
\pgfsetstrokecolor{currentstroke}%
\pgfsetdash{}{0pt}%
\pgfpathmoveto{\pgfqpoint{1.980042in}{0.647592in}}%
\pgfpathlineto{\pgfqpoint{1.980042in}{2.820000in}}%
\pgfusepath{stroke}%
\end{pgfscope}%
\begin{pgfscope}%
\pgfsetrectcap%
\pgfsetmiterjoin%
\pgfsetlinewidth{0.803000pt}%
\definecolor{currentstroke}{rgb}{0.800000,0.800000,0.800000}%
\pgfsetstrokecolor{currentstroke}%
\pgfsetdash{}{0pt}%
\pgfpathmoveto{\pgfqpoint{5.198675in}{0.647592in}}%
\pgfpathlineto{\pgfqpoint{5.198675in}{2.820000in}}%
\pgfusepath{stroke}%
\end{pgfscope}%
\begin{pgfscope}%
\pgfsetrectcap%
\pgfsetmiterjoin%
\pgfsetlinewidth{0.803000pt}%
\definecolor{currentstroke}{rgb}{0.800000,0.800000,0.800000}%
\pgfsetstrokecolor{currentstroke}%
\pgfsetdash{}{0pt}%
\pgfpathmoveto{\pgfqpoint{1.980042in}{0.647592in}}%
\pgfpathlineto{\pgfqpoint{5.198675in}{0.647592in}}%
\pgfusepath{stroke}%
\end{pgfscope}%
\begin{pgfscope}%
\pgfsetrectcap%
\pgfsetmiterjoin%
\pgfsetlinewidth{0.803000pt}%
\definecolor{currentstroke}{rgb}{0.800000,0.800000,0.800000}%
\pgfsetstrokecolor{currentstroke}%
\pgfsetdash{}{0pt}%
\pgfpathmoveto{\pgfqpoint{1.980042in}{2.820000in}}%
\pgfpathlineto{\pgfqpoint{5.198675in}{2.820000in}}%
\pgfusepath{stroke}%
\end{pgfscope}%
\begin{pgfscope}%
\pgfsetbuttcap%
\pgfsetmiterjoin%
\definecolor{currentfill}{rgb}{1.000000,1.000000,1.000000}%
\pgfsetfillcolor{currentfill}%
\pgfsetfillopacity{0.800000}%
\pgfsetlinewidth{1.003750pt}%
\definecolor{currentstroke}{rgb}{0.800000,0.800000,0.800000}%
\pgfsetstrokecolor{currentstroke}%
\pgfsetstrokeopacity{0.800000}%
\pgfsetdash{}{0pt}%
\pgfpathmoveto{\pgfqpoint{4.215519in}{2.454260in}}%
\pgfpathlineto{\pgfqpoint{5.082009in}{2.454260in}}%
\pgfpathquadraticcurveto{\pgfqpoint{5.115342in}{2.454260in}}{\pgfqpoint{5.115342in}{2.487593in}}%
\pgfpathlineto{\pgfqpoint{5.115342in}{2.703333in}}%
\pgfpathquadraticcurveto{\pgfqpoint{5.115342in}{2.736667in}}{\pgfqpoint{5.082009in}{2.736667in}}%
\pgfpathlineto{\pgfqpoint{4.215519in}{2.736667in}}%
\pgfpathquadraticcurveto{\pgfqpoint{4.182186in}{2.736667in}}{\pgfqpoint{4.182186in}{2.703333in}}%
\pgfpathlineto{\pgfqpoint{4.182186in}{2.487593in}}%
\pgfpathquadraticcurveto{\pgfqpoint{4.182186in}{2.454260in}}{\pgfqpoint{4.215519in}{2.454260in}}%
\pgfpathclose%
\pgfusepath{stroke,fill}%
\end{pgfscope}%
\begin{pgfscope}%
\pgfsetbuttcap%
\pgfsetmiterjoin%
\definecolor{currentfill}{rgb}{0.827451,0.827451,0.827451}%
\pgfsetfillcolor{currentfill}%
\pgfsetlinewidth{1.003750pt}%
\definecolor{currentstroke}{rgb}{1.000000,1.000000,1.000000}%
\pgfsetstrokecolor{currentstroke}%
\pgfsetdash{}{0pt}%
\pgfpathmoveto{\pgfqpoint{4.248852in}{2.553333in}}%
\pgfpathlineto{\pgfqpoint{4.582186in}{2.553333in}}%
\pgfpathlineto{\pgfqpoint{4.582186in}{2.670000in}}%
\pgfpathlineto{\pgfqpoint{4.248852in}{2.670000in}}%
\pgfpathclose%
\pgfusepath{stroke,fill}%
\end{pgfscope}%
\begin{pgfscope}%
\definecolor{textcolor}{rgb}{0.150000,0.150000,0.150000}%
\pgfsetstrokecolor{textcolor}%
\pgfsetfillcolor{textcolor}%
\pgftext[x=4.715519in,y=2.553333in,left,base]{\color{textcolor}\rmfamily\fontsize{12.000000}{14.400000}\selectfont ROS}%
\end{pgfscope}%
\end{pgfpicture}%
\makeatother%
\endgroup%

%% file: sections/conclusion.tex
\section{Conclusion and Outlook}
\label{sec:conclusion}
To enable remote assistance for AVs, large amounts of data must be transmitted from the vehicle to a remote operator. In order to not overload communication infrastructure, improved data compression is necessary. For this purpose, we implemented and evaluated selected state-of-the-art generative approaches for image and lidar compression, and compared them with traditional compression methods. The approaches were trained and tested with the KITTI dataset, embedded into a ROS framework and simulated within the CARLA environment. We have shown that, taking into account rate-distortion-perception-latency requirements, a real-world application can currently only be satisfied by VAEs, while GANs show the more promising results regarding reconstruction quality.

\textbf{Offline Rate-Distortion-Perception Analysis.}\label{subsec:con_offline_image_compression}
Regarding our offline rate-distortion-perception analysis, the GAN approach yields the best results for image compression and provides good reconstruction results even at very low bit-rates of 0.2 to 0.3~bpp. The reconstructions were evaluated with multiple metrics. Moreover, even at these bit-rates, it is still possible to perform object detection with a sufficient detection rate, demonstrating the perception quality of the data for remote assistance.

As we used insights gained by preliminary results of the latency evaluation during the process, we only applied a VAE for the lidar data. The reconstructions showed a mean euclidean distance error between 0.3~m and 0.5~m and especially a loss of the characteristics of lidar point clouds. Since the similarities between the point clouds are already much lower than in the image domain, we did not perform a perception analysis.

\textbf{Online Pipeline with Latency Analysis.}\label{subsec:con_online_image_compression}
Regarding the online latency evaluation, a complete end-to-end pipeline for image and lidar compression was implemented in ROS. We evaluated computing times, clearly showing that for image compression the GAN approach is significantly slower than the VAE and cannot be used for remote assistance in terms of processing speed. Thus, VAE-based compression shows a promising trade-off between processing time and reconstruction quality. For lidar compression, the bottleneck is the conversion of the point cloud to the tensor format.

\textbf{Outlook.}\label{subsec:outlook} To fully represent the domain of autonomous driving, radar data could be considered in the future. Further, the perception evaluation in the image domain could be extended to other classes such as trucks, bicycles or pedestrians. While we did perform training for image compression on a portion of the much larger WAYMO Perception Open Dataset with promising results, a complete training and evaluation on such a data set is desirable. Finally, generative video compression is a promising research area for future works.

Regarding the lidar compression, future work needs to re-evaluate the suitability of the utilized compression approach. While results such as shown in~\cite{tu_point_2019} are desirable, it must be investigated if an adapted version of their concept can be designed for the remote assistance use case.

Regarding the latency, further work will be done to optimize the inference time of the models by optimizing their structure or using acceleration methods. Since the architecture allows for processing in a single pass, a combination of camera and lidar data can be considered. Likewise, we suggest using optimized conversion methods for point clouds and ROS messages. Here, a good compromise between image quality and pipeline speed must be found. Finally, the implemented ROS framework will be deployed on our test vehicle for a real-world demonstration.

%% file: sections/acknowledgment.tex
\section{Acknowledgment}
\label{sec:acknowledgment}
This work results partly from the KIGLIS project supported by the German Federal Ministry of Education and Research (BMBF), grant number 16KIS1231.

%% file: sections/appendix.tex
\appendix

\section{Appendix}
\label{sec:appendix}
%Optionally include extra information (complete proofs, additional experiments and plots) in the appendix.
%This main paper and appendix should be submitted as a single document. There is no page limit for references or appendix.

\begin{figure*}
    \includegraphics[width=\textwidth]{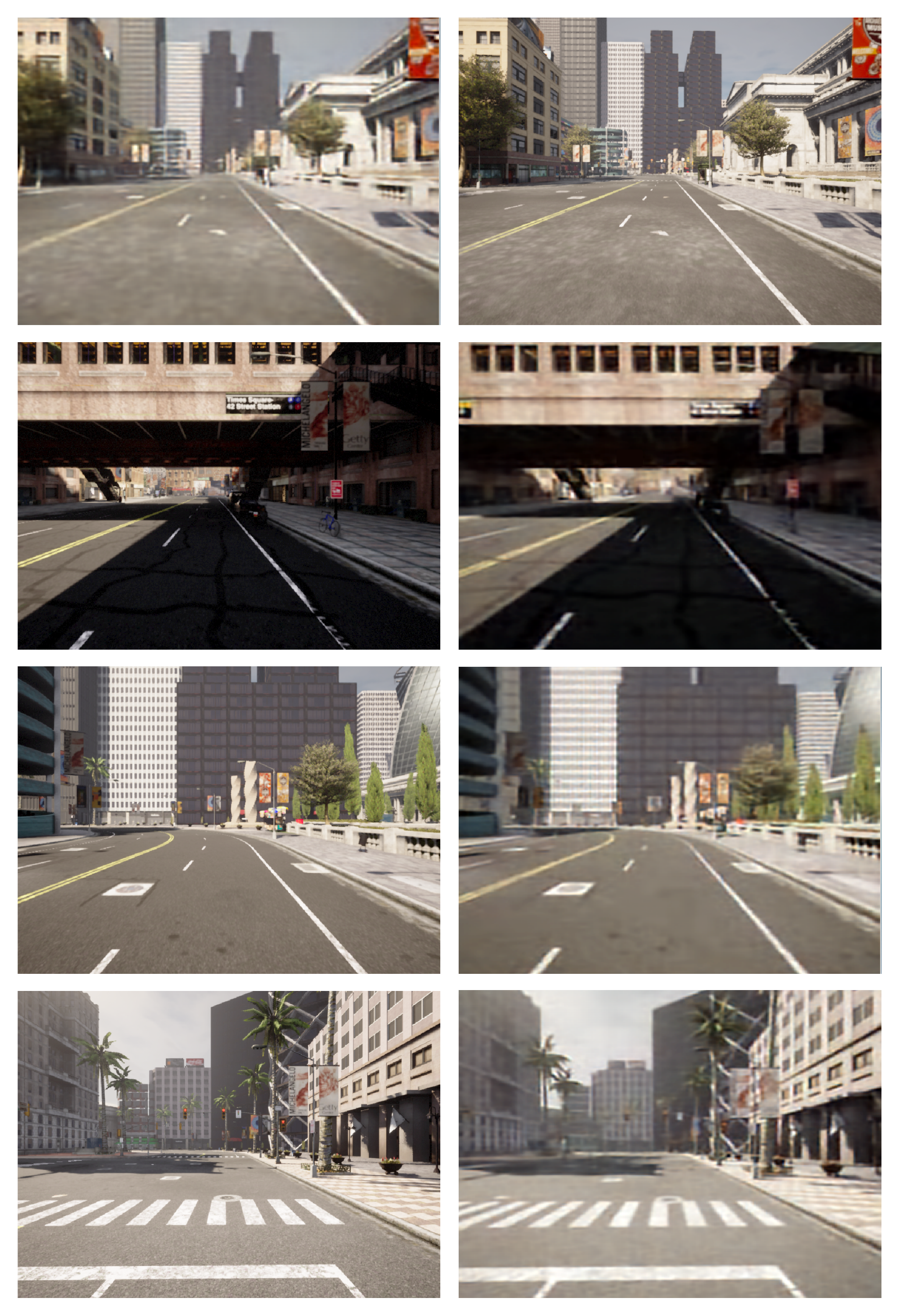}
    \caption{CARLA environment (left) and VAE-based decompressed images in RViz (right).}
\label{fig:CARLA_VAE_compression}
\end{figure*}

\begin{figure}
    \centering
    \includegraphics{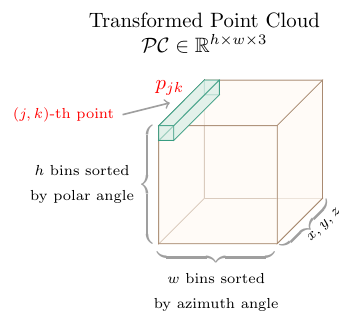}
    \caption{Sorted and binned 2d representation of the point cloud consisting of $h\times w$ bins. The $(x,y,z)$-coordinates of each point are calculated as the average of all points of the original point cloud that belong into that specific bin.}
\label{fig:point_cloud_transformed}
\end{figure}

\begin{figure}
    \begin{center}
        \input{figures/compare_cases_compression.pgf}
    \end{center}
    \caption{Out-of-domain testing of VAE trained on KITTI.}
\label{out_of_domain_figure}
\end{figure}
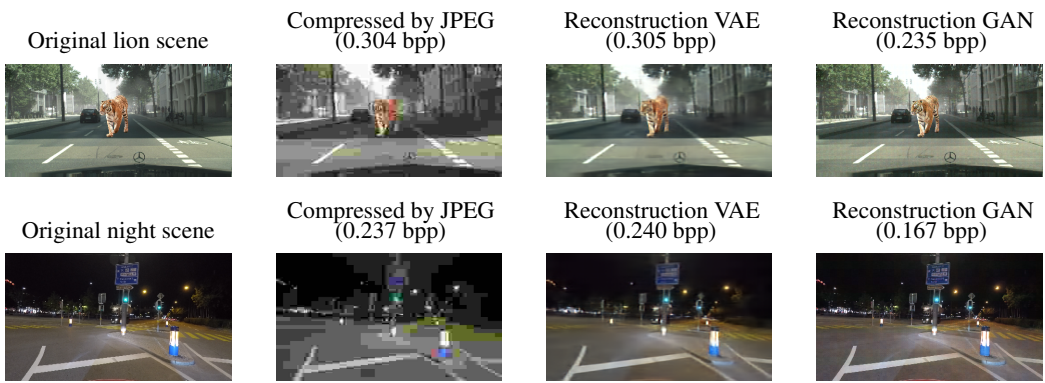

\begin{figure}
    \begin{center}
        \input{figures/appendix_images.pgf}
    \end{center}
    \caption{Examples of reconstructions with data from the KITTI dataset.}
\label{Appedix_Recons_Images}
\end{figure}
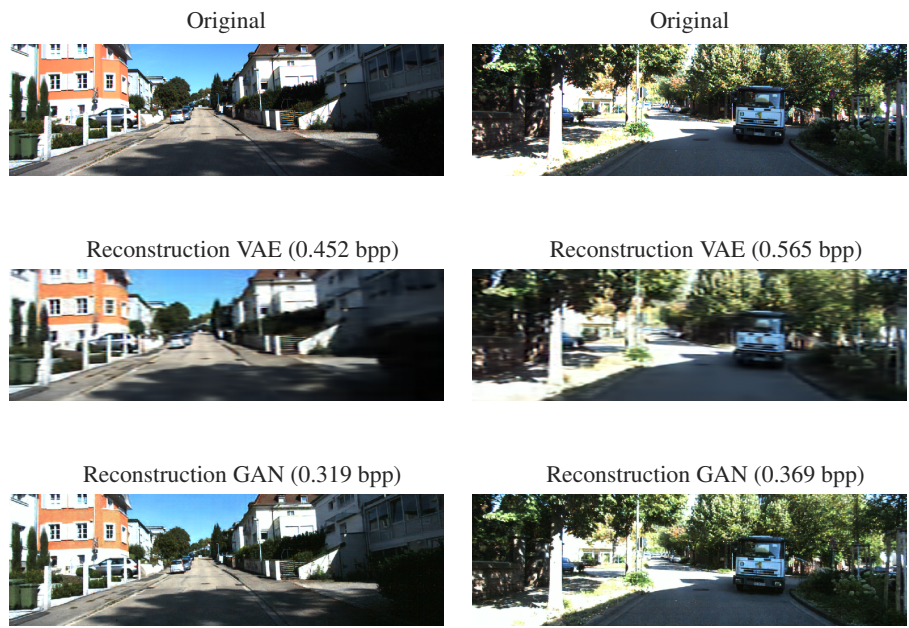

%% file: figures/compare_cases_compression.pgf
%% Creator: Matplotlib, PGF backend
%%
%% To include the figure in your LaTeX document, write
%%   \input{<filename>.pgf}
%%
%% Make sure the required packages are loaded in your preamble
%%   \usepackage{pgf}
%%
%% Figures using additional raster images can only be included by \input if
%% they are in the same directory as the main LaTeX file. For loading figures
%% from other directories you can use the `import` package
%%   \usepackage{import}
%% and then include the figures with
%%   \import{<path to file>}{<filename>.pgf}
%%
%% Matplotlib used the following preamble
%%
\begingroup%
\makeatletter%
\begin{pgfpicture}%
\pgfpathrectangle{\pgfpointorigin}{\pgfqpoint{5.625000in}{2.113327in}}%
\pgfusepath{use as bounding box, clip}%
\begin{pgfscope}%
\pgfsetbuttcap%
\pgfsetmiterjoin%
\definecolor{currentfill}{rgb}{1.000000,1.000000,1.000000}%
\pgfsetfillcolor{currentfill}%
\pgfsetlinewidth{0.000000pt}%
\definecolor{currentstroke}{rgb}{1.000000,1.000000,1.000000}%
\pgfsetstrokecolor{currentstroke}%
\pgfsetdash{}{0pt}%
\pgfpathmoveto{\pgfqpoint{0.000000in}{0.000000in}}%
\pgfpathlineto{\pgfqpoint{5.625000in}{0.000000in}}%
\pgfpathlineto{\pgfqpoint{5.625000in}{2.113327in}}%
\pgfpathlineto{\pgfqpoint{0.000000in}{2.113327in}}%
\pgfpathclose%
\pgfusepath{fill}%
\end{pgfscope}%
\begin{pgfscope}%
\pgfpathrectangle{\pgfqpoint{0.100000in}{1.166400in}}{\pgfqpoint{1.179348in}{0.589674in}}%
\pgfusepath{clip}%
\pgfsys@transformshift{0.100000in}{1.166400in}%
\pgftext[left,bottom]{\pgfimage[interpolate=true,width=1.180000in,height=0.590000in]{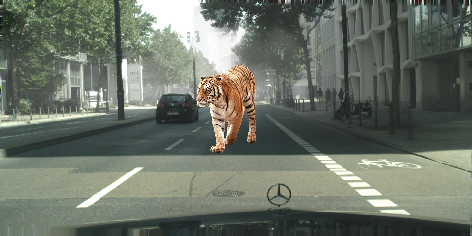}}%
\end{pgfscope}%
\begin{pgfscope}%
\definecolor{textcolor}{rgb}{0.000000,0.000000,0.000000}%
\pgfsetstrokecolor{textcolor}%
\pgfsetfillcolor{textcolor}%
\pgftext[x=0.689674in,y=1.839407in,,base]{\color{textcolor}\rmfamily\fontsize{9.000000}{8.640000}\selectfont Original lion scene}%
\end{pgfscope}%
\begin{pgfscope}%
\pgfpathrectangle{\pgfqpoint{1.515217in}{1.166400in}}{\pgfqpoint{1.179348in}{0.589674in}}%
\pgfusepath{clip}%
\pgfsys@transformshift{1.515217in}{1.166400in}%
\pgftext[left,bottom]{\pgfimage[interpolate=true,width=1.180000in,height=0.590000in]{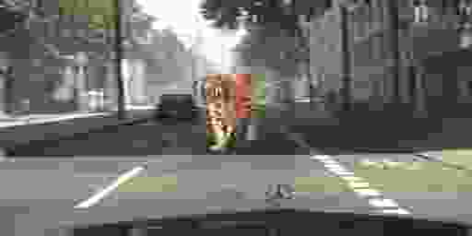}}%
\end{pgfscope}%
\begin{pgfscope}%
\definecolor{textcolor}{rgb}{0.000000,0.000000,0.000000}%
\pgfsetstrokecolor{textcolor}%
\pgfsetfillcolor{textcolor}%
\pgftext[x=1.573764in,y=1.945811in,left,base]{\color{textcolor}\rmfamily\fontsize{9.000000}{8.640000}\selectfont Compressed by JPEG}%
\end{pgfscope}%
\begin{pgfscope}%
\definecolor{textcolor}{rgb}{0.000000,0.000000,0.000000}%
\pgfsetstrokecolor{textcolor}%
\pgfsetfillcolor{textcolor}%
\pgftext[x=1.824315in,y=1.839407in,left,base]{\color{textcolor}\rmfamily\fontsize{9.000000}{8.640000}\selectfont (0.304 bpp)}%
\end{pgfscope}%
\begin{pgfscope}%
\pgfpathrectangle{\pgfqpoint{2.930435in}{1.166400in}}{\pgfqpoint{1.179348in}{0.589674in}}%
\pgfusepath{clip}%
\pgfsys@transformshift{2.930435in}{1.166400in}%
\pgftext[left,bottom]{\pgfimage[interpolate=true,width=1.180000in,height=0.590000in]{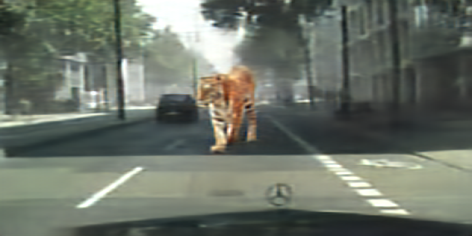}}%
\end{pgfscope}%
\begin{pgfscope}%
\definecolor{textcolor}{rgb}{0.000000,0.000000,0.000000}%
\pgfsetstrokecolor{textcolor}%
\pgfsetfillcolor{textcolor}%
\pgftext[x=3.023472in,y=1.945811in,left,base]{\color{textcolor}\rmfamily\fontsize{9.000000}{8.640000}\selectfont Reconstruction VAE}%
\end{pgfscope}%
\begin{pgfscope}%
\definecolor{textcolor}{rgb}{0.000000,0.000000,0.000000}%
\pgfsetstrokecolor{textcolor}%
\pgfsetfillcolor{textcolor}%
\pgftext[x=3.239533in,y=1.839407in,left,base]{\color{textcolor}\rmfamily\fontsize{9.000000}{8.640000}\selectfont (0.305 bpp)}%
\end{pgfscope}%
\begin{pgfscope}%
\pgfpathrectangle{\pgfqpoint{4.345652in}{1.166400in}}{\pgfqpoint{1.179348in}{0.589674in}}%
\pgfusepath{clip}%
\pgfsys@transformshift{4.345652in}{1.166400in}%
\pgftext[left,bottom]{\pgfimage[interpolate=true,width=1.180000in,height=0.590000in]{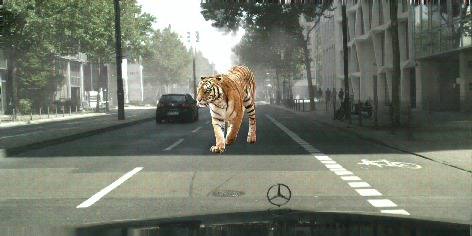}}%
\end{pgfscope}%
\begin{pgfscope}%
\definecolor{textcolor}{rgb}{0.000000,0.000000,0.000000}%
\pgfsetstrokecolor{textcolor}%
\pgfsetfillcolor{textcolor}%
\pgftext[x=4.427067in,y=1.945811in,left,base]{\color{textcolor}\rmfamily\fontsize{9.000000}{8.640000}\selectfont Reconstruction GAN}%
\end{pgfscope}%
\begin{pgfscope}%
\definecolor{textcolor}{rgb}{0.000000,0.000000,0.000000}%
\pgfsetstrokecolor{textcolor}%
\pgfsetfillcolor{textcolor}%
\pgftext[x=4.654750in,y=1.839407in,left,base]{\color{textcolor}\rmfamily\fontsize{9.000000}{8.640000}\selectfont (0.235 bpp)}%
\end{pgfscope}%
\begin{pgfscope}%
\pgfpathrectangle{\pgfqpoint{0.100000in}{0.100000in}}{\pgfqpoint{1.179348in}{0.663383in}}%
\pgfusepath{clip}%
\pgfsys@transformshift{0.100000in}{0.100000in}%
\pgftext[left,bottom]{\pgfimage[interpolate=true,width=1.180000in,height=0.665000in]{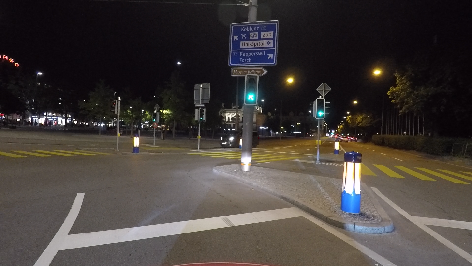}}%
\end{pgfscope}%
\begin{pgfscope}%
\definecolor{textcolor}{rgb}{0.000000,0.000000,0.000000}%
\pgfsetstrokecolor{textcolor}%
\pgfsetfillcolor{textcolor}%
\pgftext[x=0.689674in,y=0.846716in,,base]{\color{textcolor}\rmfamily\fontsize{9.000000}{8.640000}\selectfont Original night scene}%
\end{pgfscope}%
\begin{pgfscope}%
\pgfpathrectangle{\pgfqpoint{1.515217in}{0.100000in}}{\pgfqpoint{1.179348in}{0.663383in}}%
\pgfusepath{clip}%
\pgfsys@transformshift{1.515217in}{0.100000in}%
\pgftext[left,bottom]{\pgfimage[interpolate=true,width=1.180000in,height=0.665000in]{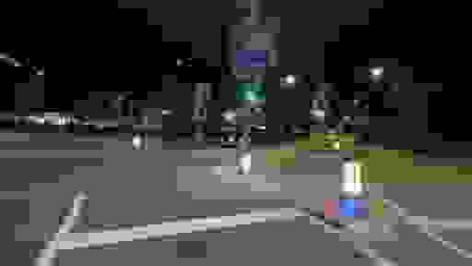}}%
\end{pgfscope}%
\begin{pgfscope}%
\definecolor{textcolor}{rgb}{0.000000,0.000000,0.000000}%
\pgfsetstrokecolor{textcolor}%
\pgfsetfillcolor{textcolor}%
\pgftext[x=1.573764in,y=0.953121in,left,base]{\color{textcolor}\rmfamily\fontsize{9.000000}{8.640000}\selectfont Compressed by JPEG}%
\end{pgfscope}%
\begin{pgfscope}%
\definecolor{textcolor}{rgb}{0.000000,0.000000,0.000000}%
\pgfsetstrokecolor{textcolor}%
\pgfsetfillcolor{textcolor}%
\pgftext[x=1.824315in,y=0.846716in,left,base]{\color{textcolor}\rmfamily\fontsize{9.000000}{8.640000}\selectfont (0.237 bpp)}%
\end{pgfscope}%
\begin{pgfscope}%
\pgfpathrectangle{\pgfqpoint{2.930435in}{0.100000in}}{\pgfqpoint{1.179348in}{0.663383in}}%
\pgfusepath{clip}%
\pgfsys@transformshift{2.930435in}{0.100000in}%
\pgftext[left,bottom]{\pgfimage[interpolate=true,width=1.180000in,height=0.665000in]{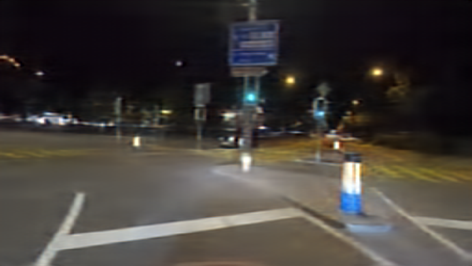}}%
\end{pgfscope}%
\begin{pgfscope}%
\definecolor{textcolor}{rgb}{0.000000,0.000000,0.000000}%
\pgfsetstrokecolor{textcolor}%
\pgfsetfillcolor{textcolor}%
\pgftext[x=3.023472in,y=0.953121in,left,base]{\color{textcolor}\rmfamily\fontsize{9.000000}{8.640000}\selectfont Reconstruction VAE}%
\end{pgfscope}%
\begin{pgfscope}%
\definecolor{textcolor}{rgb}{0.000000,0.000000,0.000000}%
\pgfsetstrokecolor{textcolor}%
\pgfsetfillcolor{textcolor}%
\pgftext[x=3.239533in,y=0.846716in,left,base]{\color{textcolor}\rmfamily\fontsize{9.000000}{8.640000}\selectfont (0.240 bpp)}%
\end{pgfscope}%
\begin{pgfscope}%
\pgfpathrectangle{\pgfqpoint{4.345652in}{0.100000in}}{\pgfqpoint{1.179348in}{0.663383in}}%
\pgfusepath{clip}%
\pgfsys@transformshift{4.345652in}{0.100000in}%
\pgftext[left,bottom]{\pgfimage[interpolate=true,width=1.180000in,height=0.665000in]{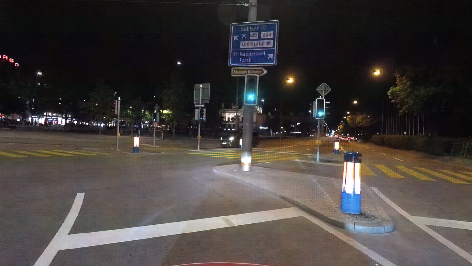}}%
\end{pgfscope}%
\begin{pgfscope}%
\definecolor{textcolor}{rgb}{0.000000,0.000000,0.000000}%
\pgfsetstrokecolor{textcolor}%
\pgfsetfillcolor{textcolor}%
\pgftext[x=4.427067in,y=0.953121in,left,base]{\color{textcolor}\rmfamily\fontsize{9.000000}{8.640000}\selectfont Reconstruction GAN}%
\end{pgfscope}%
\begin{pgfscope}%
\definecolor{textcolor}{rgb}{0.000000,0.000000,0.000000}%
\pgfsetstrokecolor{textcolor}%
\pgfsetfillcolor{textcolor}%
\pgftext[x=4.654750in,y=0.846716in,left,base]{\color{textcolor}\rmfamily\fontsize{9.000000}{8.640000}\selectfont (0.167 bpp)}%
\end{pgfscope}%
\end{pgfpicture}%
\makeatother%
\endgroup%

%% file: figures/appendix_images.pgf
%% Creator: Matplotlib, PGF backend
%%
%% To include the figure in your LaTeX document, write
%%   \input{<filename>.pgf}
%%
%% Make sure the required packages are loaded in your preamble
%%   \usepackage{pgf}
%%
%% Figures using additional raster images can only be included by \input if
%% they are in the same directory as the main LaTeX file. For loading figures
%% from other directories you can use the `import` package
%%   \usepackage{import}
%% and then include the figures with
%%   \import{<path to file>}{<filename>.pgf}
%%
%% Matplotlib used the following preamble
%%
\begingroup%
\makeatletter%
\begin{pgfpicture}%
\pgfpathrectangle{\pgfpointorigin}{\pgfqpoint{4.900000in}{3.441646in}}%
\pgfusepath{use as bounding box, clip}%
\begin{pgfscope}%
\pgfsetbuttcap%
\pgfsetmiterjoin%
\definecolor{currentfill}{rgb}{1.000000,1.000000,1.000000}%
\pgfsetfillcolor{currentfill}%
\pgfsetlinewidth{0.000000pt}%
\definecolor{currentstroke}{rgb}{1.000000,1.000000,1.000000}%
\pgfsetstrokecolor{currentstroke}%
\pgfsetdash{}{0pt}%
\pgfpathclose%
\pgfusepath{fill}%
\end{pgfscope}%
\begin{pgfscope}%
\pgfsetbuttcap%
\pgfsetmiterjoin%
\definecolor{currentfill}{rgb}{1.000000,1.000000,1.000000}%
\pgfsetfillcolor{currentfill}%
\pgfsetlinewidth{0.000000pt}%
\definecolor{currentstroke}{rgb}{0.000000,0.000000,0.000000}%
\pgfsetstrokecolor{currentstroke}%
\pgfsetstrokeopacity{0.000000}%
\pgfsetdash{}{0pt}%
\pgfpathclose%
\pgfusepath{fill}%
\end{pgfscope}%
\begin{pgfscope}%
\pgfpathrectangle{\pgfqpoint{0.100000in}{2.455676in}}{\pgfqpoint{2.275000in}{0.686896in}}%
\pgfusepath{clip}%
\pgfsys@transformshift{0.100000in}{2.455676in}%
\pgftext[left,bottom]{\pgfimage[interpolate=true,width=2.275000in,height=0.687500in]{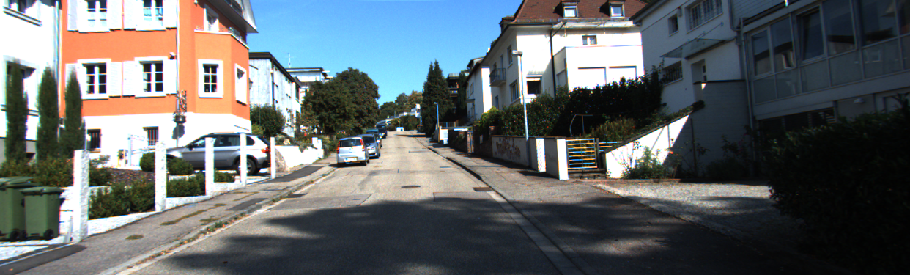}}%
\end{pgfscope}%
\begin{pgfscope}%
\pgfsetrectcap%
\pgfsetmiterjoin%
\pgfsetlinewidth{1.254687pt}%
\definecolor{currentstroke}{rgb}{0.150000,0.150000,0.150000}%
\pgfsetstrokecolor{currentstroke}%
\pgfsetdash{}{0pt}%
\pgfusepath{stroke}%
\end{pgfscope}%
\begin{pgfscope}%
\pgfsetrectcap%
\pgfsetmiterjoin%
\pgfsetlinewidth{1.254687pt}%
\definecolor{currentstroke}{rgb}{0.150000,0.150000,0.150000}%
\pgfsetstrokecolor{currentstroke}%
\pgfsetdash{}{0pt}%
\pgfusepath{stroke}%
\end{pgfscope}%
\begin{pgfscope}%
\pgfsetrectcap%
\pgfsetmiterjoin%
\pgfsetlinewidth{1.254687pt}%
\definecolor{currentstroke}{rgb}{0.150000,0.150000,0.150000}%
\pgfsetstrokecolor{currentstroke}%
\pgfsetdash{}{0pt}%
\pgfusepath{stroke}%
\end{pgfscope}%
\begin{pgfscope}%
\pgfsetrectcap%
\pgfsetmiterjoin%
\pgfsetlinewidth{1.254687pt}%
\definecolor{currentstroke}{rgb}{0.150000,0.150000,0.150000}%
\pgfsetstrokecolor{currentstroke}%
\pgfsetdash{}{0pt}%
\pgfusepath{stroke}%
\end{pgfscope}%
\begin{pgfscope}%
\definecolor{textcolor}{rgb}{0.150000,0.150000,0.150000}%
\pgfsetstrokecolor{textcolor}%
\pgfsetfillcolor{textcolor}%
\pgftext[x=1.237500in,y=3.225906in,,base]{\color{textcolor}\rmfamily\fontsize{9.000000}{14.400000}\selectfont Original}%
\end{pgfscope}%
\begin{pgfscope}%
\pgfsetbuttcap%
\pgfsetmiterjoin%
\definecolor{currentfill}{rgb}{1.000000,1.000000,1.000000}%
\pgfsetfillcolor{currentfill}%
\pgfsetlinewidth{0.000000pt}%
\definecolor{currentstroke}{rgb}{0.000000,0.000000,0.000000}%
\pgfsetstrokecolor{currentstroke}%
\pgfsetstrokeopacity{0.000000}%
\pgfsetdash{}{0pt}%
\pgfpathclose%
\pgfusepath{fill}%
\end{pgfscope}%
\begin{pgfscope}%
\pgfpathrectangle{\pgfqpoint{2.525000in}{2.455676in}}{\pgfqpoint{2.275000in}{0.686896in}}%
\pgfusepath{clip}%
\pgfsys@transformshift{2.525000in}{2.455676in}%
\pgftext[left,bottom]{\pgfimage[interpolate=true,width=2.275000in,height=0.687500in]{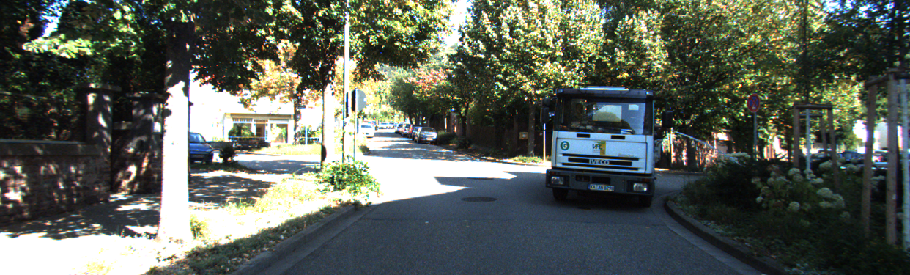}}%
\end{pgfscope}%
\begin{pgfscope}%
\pgfsetrectcap%
\pgfsetmiterjoin%
\pgfsetlinewidth{1.254687pt}%
\definecolor{currentstroke}{rgb}{0.150000,0.150000,0.150000}%
\pgfsetstrokecolor{currentstroke}%
\pgfsetdash{}{0pt}%
\pgfusepath{stroke}%
\end{pgfscope}%
\begin{pgfscope}%
\pgfsetrectcap%
\pgfsetmiterjoin%
\pgfsetlinewidth{1.254687pt}%
\definecolor{currentstroke}{rgb}{0.150000,0.150000,0.150000}%
\pgfsetstrokecolor{currentstroke}%
\pgfsetdash{}{0pt}%
\pgfusepath{stroke}%
\end{pgfscope}%
\begin{pgfscope}%
\pgfsetrectcap%
\pgfsetmiterjoin%
\pgfsetlinewidth{1.254687pt}%
\definecolor{currentstroke}{rgb}{0.150000,0.150000,0.150000}%
\pgfsetstrokecolor{currentstroke}%
\pgfsetdash{}{0pt}%
\pgfusepath{stroke}%
\end{pgfscope}%
\begin{pgfscope}%
\pgfsetrectcap%
\pgfsetmiterjoin%
\pgfsetlinewidth{1.254687pt}%
\definecolor{currentstroke}{rgb}{0.150000,0.150000,0.150000}%
\pgfsetstrokecolor{currentstroke}%
\pgfsetdash{}{0pt}%
\pgfusepath{stroke}%
\end{pgfscope}%
\begin{pgfscope}%
\definecolor{textcolor}{rgb}{0.150000,0.150000,0.150000}%
\pgfsetstrokecolor{textcolor}%
\pgfsetfillcolor{textcolor}%
\pgftext[x=3.662500in,y=3.225906in,,base]{\color{textcolor}\rmfamily\fontsize{9.000000}{14.400000}\selectfont Original}%
\end{pgfscope}%
\begin{pgfscope}%
\pgfsetbuttcap%
\pgfsetmiterjoin%
\definecolor{currentfill}{rgb}{1.000000,1.000000,1.000000}%
\pgfsetfillcolor{currentfill}%
\pgfsetlinewidth{0.000000pt}%
\definecolor{currentstroke}{rgb}{0.000000,0.000000,0.000000}%
\pgfsetstrokecolor{currentstroke}%
\pgfsetstrokeopacity{0.000000}%
\pgfsetdash{}{0pt}%
\pgfpathclose%
\pgfusepath{fill}%
\end{pgfscope}%
\begin{pgfscope}%
\pgfpathrectangle{\pgfqpoint{0.100000in}{1.277838in}}{\pgfqpoint{2.275000in}{0.686896in}}%
\pgfusepath{clip}%
\pgfsys@transformshift{0.100000in}{1.277838in}%
\pgftext[left,bottom]{\pgfimage[interpolate=true,width=2.275000in,height=0.687500in]{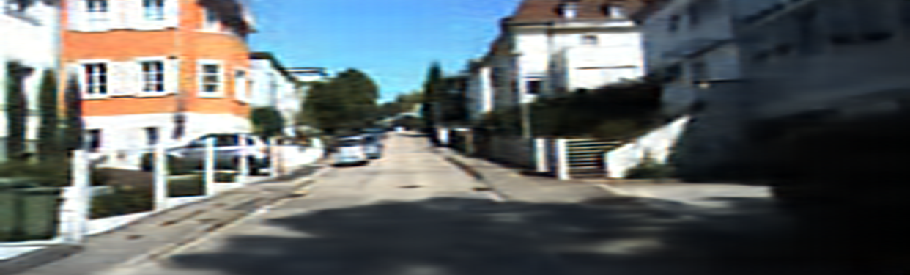}}%
\end{pgfscope}%
\begin{pgfscope}%
\pgfsetrectcap%
\pgfsetmiterjoin%
\pgfsetlinewidth{1.254687pt}%
\definecolor{currentstroke}{rgb}{0.150000,0.150000,0.150000}%
\pgfsetstrokecolor{currentstroke}%
\pgfsetdash{}{0pt}%
\pgfusepath{stroke}%
\end{pgfscope}%
\begin{pgfscope}%
\pgfsetrectcap%
\pgfsetmiterjoin%
\pgfsetlinewidth{1.254687pt}%
\definecolor{currentstroke}{rgb}{0.150000,0.150000,0.150000}%
\pgfsetstrokecolor{currentstroke}%
\pgfsetdash{}{0pt}%
\pgfusepath{stroke}%
\end{pgfscope}%
\begin{pgfscope}%
\pgfsetrectcap%
\pgfsetmiterjoin%
\pgfsetlinewidth{1.254687pt}%
\definecolor{currentstroke}{rgb}{0.150000,0.150000,0.150000}%
\pgfsetstrokecolor{currentstroke}%
\pgfsetdash{}{0pt}%
\pgfusepath{stroke}%
\end{pgfscope}%
\begin{pgfscope}%
\pgfsetrectcap%
\pgfsetmiterjoin%
\pgfsetlinewidth{1.254687pt}%
\definecolor{currentstroke}{rgb}{0.150000,0.150000,0.150000}%
\pgfsetstrokecolor{currentstroke}%
\pgfsetdash{}{0pt}%
\pgfusepath{stroke}%
\end{pgfscope}%
\begin{pgfscope}%
\definecolor{textcolor}{rgb}{0.150000,0.150000,0.150000}%
\pgfsetstrokecolor{textcolor}%
\pgfsetfillcolor{textcolor}%
\pgftext[x=0.505033in,y=2.030475in,left,base]{\color{textcolor}\rmfamily\fontsize{9.000000}{14.400000}\selectfont Reconstruction VAE (0.452 bpp)}%
\end{pgfscope}%
\begin{pgfscope}%
\pgfsetbuttcap%
\pgfsetmiterjoin%
\definecolor{currentfill}{rgb}{1.000000,1.000000,1.000000}%
\pgfsetfillcolor{currentfill}%
\pgfsetlinewidth{0.000000pt}%
\definecolor{currentstroke}{rgb}{0.000000,0.000000,0.000000}%
\pgfsetstrokecolor{currentstroke}%
\pgfsetstrokeopacity{0.000000}%
\pgfsetdash{}{0pt}%
\pgfpathclose%
\pgfusepath{fill}%
\end{pgfscope}%
\begin{pgfscope}%
\pgfpathrectangle{\pgfqpoint{2.525000in}{1.277838in}}{\pgfqpoint{2.275000in}{0.686896in}}%
\pgfusepath{clip}%
\pgfsys@transformshift{2.525000in}{1.277838in}%
\pgftext[left,bottom]{\pgfimage[interpolate=true,width=2.275000in,height=0.687500in]{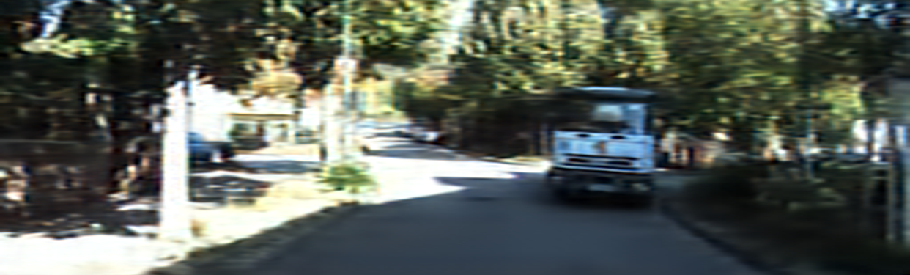}}%
\end{pgfscope}%
\begin{pgfscope}%
\pgfsetrectcap%
\pgfsetmiterjoin%
\pgfsetlinewidth{1.254687pt}%
\definecolor{currentstroke}{rgb}{0.150000,0.150000,0.150000}%
\pgfsetstrokecolor{currentstroke}%
\pgfsetdash{}{0pt}%
\pgfusepath{stroke}%
\end{pgfscope}%
\begin{pgfscope}%
\pgfsetrectcap%
\pgfsetmiterjoin%
\pgfsetlinewidth{1.254687pt}%
\definecolor{currentstroke}{rgb}{0.150000,0.150000,0.150000}%
\pgfsetstrokecolor{currentstroke}%
\pgfsetdash{}{0pt}%
\pgfusepath{stroke}%
\end{pgfscope}%
\begin{pgfscope}%
\pgfsetrectcap%
\pgfsetmiterjoin%
\pgfsetlinewidth{1.254687pt}%
\definecolor{currentstroke}{rgb}{0.150000,0.150000,0.150000}%
\pgfsetstrokecolor{currentstroke}%
\pgfsetdash{}{0pt}%
\pgfusepath{stroke}%
\end{pgfscope}%
\begin{pgfscope}%
\pgfsetrectcap%
\pgfsetmiterjoin%
\pgfsetlinewidth{1.254687pt}%
\definecolor{currentstroke}{rgb}{0.150000,0.150000,0.150000}%
\pgfsetstrokecolor{currentstroke}%
\pgfsetdash{}{0pt}%
\pgfusepath{stroke}%
\end{pgfscope}%
\begin{pgfscope}%
\definecolor{textcolor}{rgb}{0.150000,0.150000,0.150000}%
\pgfsetstrokecolor{textcolor}%
\pgfsetfillcolor{textcolor}%
\pgftext[x=2.930033in,y=2.030475in,left,base]{\color{textcolor}\rmfamily\fontsize{9.000000}{14.400000}\selectfont Reconstruction VAE (0.565 bpp)}%
\end{pgfscope}%
\begin{pgfscope}%
\pgfsetbuttcap%
\pgfsetmiterjoin%
\definecolor{currentfill}{rgb}{1.000000,1.000000,1.000000}%
\pgfsetfillcolor{currentfill}%
\pgfsetlinewidth{0.000000pt}%
\definecolor{currentstroke}{rgb}{0.000000,0.000000,0.000000}%
\pgfsetstrokecolor{currentstroke}%
\pgfsetstrokeopacity{0.000000}%
\pgfsetdash{}{0pt}%
\pgfpathclose%
\pgfusepath{fill}%
\end{pgfscope}%
\begin{pgfscope}%
\pgfpathrectangle{\pgfqpoint{0.100000in}{0.100000in}}{\pgfqpoint{2.275000in}{0.686896in}}%
\pgfusepath{clip}%
\pgfsys@transformshift{0.100000in}{0.100000in}%
\pgftext[left,bottom]{\pgfimage[interpolate=true,width=2.275000in,height=0.687500in]{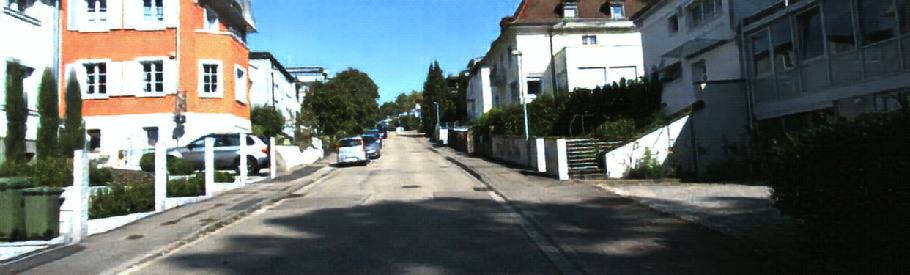}}%
\end{pgfscope}%
\begin{pgfscope}%
\pgfsetrectcap%
\pgfsetmiterjoin%
\pgfsetlinewidth{1.254687pt}%
\definecolor{currentstroke}{rgb}{0.150000,0.150000,0.150000}%
\pgfsetstrokecolor{currentstroke}%
\pgfsetdash{}{0pt}%
\pgfusepath{stroke}%
\end{pgfscope}%
\begin{pgfscope}%
\pgfsetrectcap%
\pgfsetmiterjoin%
\pgfsetlinewidth{1.254687pt}%
\definecolor{currentstroke}{rgb}{0.150000,0.150000,0.150000}%
\pgfsetstrokecolor{currentstroke}%
\pgfsetdash{}{0pt}%
\pgfusepath{stroke}%
\end{pgfscope}%
\begin{pgfscope}%
\pgfsetrectcap%
\pgfsetmiterjoin%
\pgfsetlinewidth{1.254687pt}%
\definecolor{currentstroke}{rgb}{0.150000,0.150000,0.150000}%
\pgfsetstrokecolor{currentstroke}%
\pgfsetdash{}{0pt}%
\pgfusepath{stroke}%
\end{pgfscope}%
\begin{pgfscope}%
\pgfsetrectcap%
\pgfsetmiterjoin%
\pgfsetlinewidth{1.254687pt}%
\definecolor{currentstroke}{rgb}{0.150000,0.150000,0.150000}%
\pgfsetstrokecolor{currentstroke}%
\pgfsetdash{}{0pt}%
\pgfusepath{stroke}%
\end{pgfscope}%
\begin{pgfscope}%
\definecolor{textcolor}{rgb}{0.150000,0.150000,0.150000}%
\pgfsetstrokecolor{textcolor}%
\pgfsetfillcolor{textcolor}%
\pgftext[x=0.487459in,y=0.852637in,left,base]{\color{textcolor}\rmfamily\fontsize{9.000000}{14.400000}\selectfont Reconstruction GAN (0.319 bpp)}%
\end{pgfscope}%
\begin{pgfscope}%
\pgfsetbuttcap%
\pgfsetmiterjoin%
\definecolor{currentfill}{rgb}{1.000000,1.000000,1.000000}%
\pgfsetfillcolor{currentfill}%
\pgfsetlinewidth{0.000000pt}%
\definecolor{currentstroke}{rgb}{0.000000,0.000000,0.000000}%
\pgfsetstrokecolor{currentstroke}%
\pgfsetstrokeopacity{0.000000}%
\pgfsetdash{}{0pt}%
\pgfpathclose%
\pgfusepath{fill}%
\end{pgfscope}%
\begin{pgfscope}%
\pgfpathrectangle{\pgfqpoint{2.525000in}{0.100000in}}{\pgfqpoint{2.275000in}{0.686896in}}%
\pgfusepath{clip}%
\pgfsys@transformshift{2.525000in}{0.100000in}%
\pgftext[left,bottom]{\pgfimage[interpolate=true,width=2.275000in,height=0.687500in]{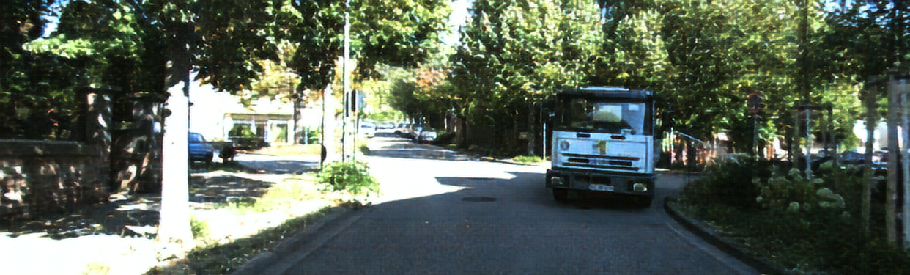}}%
\end{pgfscope}%
\begin{pgfscope}%
\pgfsetrectcap%
\pgfsetmiterjoin%
\pgfsetlinewidth{1.254687pt}%
\definecolor{currentstroke}{rgb}{0.150000,0.150000,0.150000}%
\pgfsetstrokecolor{currentstroke}%
\pgfsetdash{}{0pt}%
\pgfusepath{stroke}%
\end{pgfscope}%
\begin{pgfscope}%
\pgfsetrectcap%
\pgfsetmiterjoin%
\pgfsetlinewidth{1.254687pt}%
\definecolor{currentstroke}{rgb}{0.150000,0.150000,0.150000}%
\pgfsetstrokecolor{currentstroke}%
\pgfsetdash{}{0pt}%
\pgfusepath{stroke}%
\end{pgfscope}%
\begin{pgfscope}%
\pgfsetrectcap%
\pgfsetmiterjoin%
\pgfsetlinewidth{1.254687pt}%
\definecolor{currentstroke}{rgb}{0.150000,0.150000,0.150000}%
\pgfsetstrokecolor{currentstroke}%
\pgfsetdash{}{0pt}%
\pgfusepath{stroke}%
\end{pgfscope}%
\begin{pgfscope}%
\pgfsetrectcap%
\pgfsetmiterjoin%
\pgfsetlinewidth{1.254687pt}%
\definecolor{currentstroke}{rgb}{0.150000,0.150000,0.150000}%
\pgfsetstrokecolor{currentstroke}%
\pgfsetdash{}{0pt}%
\pgfusepath{stroke}%
\end{pgfscope}%
\begin{pgfscope}%
\definecolor{textcolor}{rgb}{0.150000,0.150000,0.150000}%
\pgfsetstrokecolor{textcolor}%
\pgfsetfillcolor{textcolor}%
\pgftext[x=2.912459in,y=0.852637in,left,base]{\color{textcolor}\rmfamily\fontsize{9.000000}{14.400000}\selectfont Reconstruction GAN (0.369 bpp)}%
\end{pgfscope}%
\end{pgfpicture}%
\makeatother%
\endgroup%